%% file: main.tex
\documentclass[11pt]{article}

\usepackage[preprint]{acl}

\usepackage{times}
\usepackage{latexsym}
\usepackage[T1]{fontenc}
\usepackage[utf8]{inputenc}
\usepackage{microtype}
\usepackage{inconsolata}
\usepackage{graphicx}
\usepackage{booktabs}   % generated tables use \toprule/\midrule/\bottomrule
\usepackage{tabularx}   % T6 has prose cells; X columns absorb slack, never overflow
\usepackage{amsmath}

\graphicspath{{figures/}}

\input{macros/numbers}
\title{Business Truth, not SQL Accuracy: A Rule-Gated 7B Analytics\\
       Agent Outperforms a Direct-Prompted 32B Baseline}

\author{Morris Lee \\
  Independent Researcher \\
  \url{https://github.com/k-w-lee/query_proof} \\}

\begin{document}
\maketitle

\begin{abstract}
% Written LAST, after 06_results and 09_conclusion. 150-250 words.
% Must contain: problem, gap, system, benchmark, method, main result, and at
% least one negative result (docs/prd-publication.md section 8).
\input{sections/00_abstract}
\end{abstract}

\input{sections/01_introduction}
\input{sections/02_related_work}
\input{sections/03_benchmark}
\input{sections/04_method}
\input{sections/05_setup}
\input{sections/06_results}
\input{sections/08_limitations}
\input{sections/09_conclusion}

% acl.sty already issues \bibliographystyle{acl_natbib}; setting it again makes
% BibTeX abort with "Illegal, another \bibstyle command".
\bibliography{references}

\appendix
\input{appendix/a_scoring_corrections}

\input{appendix/b_reproduction}
\input{appendix/c_supporting_floats}

\input{appendix/d_gate_a}
\input{appendix/e_metrics}

\end{document}

%% file: macros/numbers.tex
\newcommand{\wrbVersion}{0.2.0}                                   % benchmark_version
\newcommand{\nTasksTotal}{400}                                    % composition.{dev,validation,test}.n
\newcommand{\nDevSplit}{240}                                      % composition.dev.n
\newcommand{\nValidationSplit}{80}                                % composition.validation.n
\newcommand{\nTestSplit}{80}                                      % composition.test.n
\newcommand{\adjudicated}{50}                                     % adjudication.n_adjudicated
\newcommand{\adjudicationAgreement}{0.920}                        % adjudication.raw_agreement
\newcommand{\adjudicationKappa}{0.887}                            % adjudication.cohens_kappa
\newcommand{\adjudicationMeasure}{test--retest}                   % adjudication.measure
\newcommand{\runsCommit}{3fa9b5d}                                 % provenance.runs_commit
\newcommand{\analysisCommit}{7116b07}                             % provenance.analysis_commit
\newcommand{\nHashedInputs}{26}                                   % provenance.inputs
\newcommand{\nTestTasks}{80}                                      % splits.test.n_tasks
\newcommand{\btrRoutedTest}{0.537}                                % splits.test.systems.QueryProof routed.success
\newcommand{\fsrRoutedTest}{0.351}                                % splits.test.systems.QueryProof routed.false_success
\newcommand{\covRoutedTest}{0.600}                                % splits.test.systems.QueryProof routed.coverage
\newcommand{\ansaccRoutedTest}{1.000}                             % splits.test.systems.QueryProof routed.answer_accuracy
\newcommand{\advRoutedTest}{1.000}                                % splits.test.systems.QueryProof routed.adversarial_pass
\newcommand{\btrCIRoutedTest}{$[0.425, 0.638]$}                   % splits.test.systems.QueryProof routed.ci
\newcommand{\unsafeRoutedTest}{0}                                 % splits.test.systems.QueryProof routed.by_category.*.unsafe_attempts
\newcommand{\fsrCountRoutedTest}{13}                              % splits.test.systems.QueryProof routed.false_success_count
\newcommand{\btrDropRouted}{$-0.200$}                             % splits.test.systems.QueryProof routed.success - splits.validation.systems.QueryProof routed.success
\newcommand{\btrBaseTest}{0.562}                                  % splits.test.systems.QueryProof base.success
\newcommand{\covBaseTest}{0.700}                                  % splits.test.systems.QueryProof base.coverage
\newcommand{\ansaccBaseTest}{0.929}                               % splits.test.systems.QueryProof base.answer_accuracy
\newcommand{\btrCIBaseTest}{$[0.450, 0.675]$}                     % splits.test.systems.QueryProof base.ci
\newcommand{\btrThirtyTwoBTest}{0.300}                            % splits.test.systems.32B direct.success
\newcommand{\fsrThirtyTwoBTest}{0.754}                            % splits.test.systems.32B direct.false_success
\newcommand{\btrCIThirtyTwoBTest}{$[0.200, 0.400]$}               % splits.test.systems.32B direct.ci
\newcommand{\btrDropThirtyTwoB}{$-0.175$}                         % splits.test.systems.32B direct.success - splits.validation.systems.32B direct.success
\newcommand{\btrDropSevenBRetrieval}{$+0.050$}                    % splits.test.systems.7B retrieval.success - splits.validation.systems.7B retrieval.success
\newcommand{\btrSevenBFewShotTest}{0.275}                         % splits.test.systems.7B few-shot.success
\newcommand{\fsrSevenBFewShotTest}{0.797}                         % splits.test.systems.7B few-shot.false_success
\newcommand{\covSevenBFewShotTest}{1.000}                         % splits.test.systems.7B few-shot.coverage
\newcommand{\ansaccSevenBFewShotTest}{0.350}                      % splits.test.systems.7B few-shot.answer_accuracy
\newcommand{\btrCISevenBFewShotTest}{$[0.175, 0.375]$}            % splits.test.systems.7B few-shot.ci
\newcommand{\fsrSevenBDirectTest}{1.000}                          % splits.test.systems.7B direct.false_success
\newcommand{\btrRoutedVal}{0.738}                                 % splits.validation.systems.QueryProof routed.success
\newcommand{\btrBaseVal}{0.725}                                   % splits.validation.systems.QueryProof base.success
\newcommand{\btrRoutedStandard}{0.500}                            % splits.test.systems.QueryProof routed.by_category.standard.success_rate
\newcommand{\btrRoutedUnanswerable}{0.438}                        % splits.test.systems.QueryProof routed.by_category.unanswerable.success_rate
\newcommand{\btrRoutedDrift}{1.000}                               % splits.test.systems.QueryProof routed.by_category.schema_drift.success_rate
\newcommand{\btrThirtyTwoBStandard}{0.500}                        % splits.test.systems.32B direct.by_category.standard.success_rate
\newcommand{\btrThirtyTwoBUnanswerable}{0.000}                    % splits.test.systems.32B direct.by_category.unanswerable.success_rate
\newcommand{\btrThirtyTwoBDrift}{0.000}                           % splits.test.systems.32B direct.by_category.schema_drift.success_rate
\newcommand{\nTestStandard}{32}                                   % composition.test.by_category_domain.standard
\newcommand{\nTestAmbiguous}{16}                                  % composition.test.by_category_domain.ambiguous
\newcommand{\nTestUnanswerable}{16}                               % composition.test.by_category_domain.unanswerable
\newcommand{\nTestDrift}{8}                                       % composition.test.by_category_domain.schema_drift
\newcommand{\nTestAdversarial}{8}                                 % composition.test.by_category_domain.adversarial
\newcommand{\pairedResamplesTest}{2000}                           % paired.test.comparisons[0].metrics.success.resamples
\newcommand{\deltaBtrThirtyTwoBTest}{$+0.237$}                    % paired.test.comparisons[qwen25coder32b_direct].success.difference
\newcommand{\deltaBtrThirtyTwoBTestCI}{$[+0.112, +0.375]$}        % paired.test.comparisons[qwen25coder32b_direct].success.ci_95
\newcommand{\verbBtrThirtyTwoBTest}{outperforms}                  % paired.test.comparisons[qwen25coder32b_direct].success.permitted_verb
\newcommand{\discordantBtrThirtyTwoBTest}{27/8/45}                % paired.test.comparisons[qwen25coder32b_direct].success a_better/b_better/tied
\newcommand{\deltaFsShareThirtyTwoBTest}{$-0.450$}                % paired.test.comparisons[qwen25coder32b_direct].false_success_share.difference
\newcommand{\deltaFsShareThirtyTwoBTestCI}{$[-0.575, -0.337]$}    % paired.test.comparisons[qwen25coder32b_direct].false_success_share.ci_95
\newcommand{\clusterBtrThirtyTwoBTestCI}{$[-0.125, +0.562]$}      % paired.test.comparisons[qwen25coder32b_direct].success_cluster.ci_95
\newcommand{\nFamiliesTest}{10}                                   % paired.test.comparisons[qwen25coder32b_direct].success_cluster.n_families
\newcommand{\deltaCpcaThirtyTwoBTest}{$-0.00414$}                 % paired.test.comparisons[qwen25coder32b_direct].cpca.difference
\newcommand{\deltaCpcaThirtyTwoBTestCI}{$[-0.00752, -0.00250]$}   % paired.test.comparisons[qwen25coder32b_direct].cpca.ci_95
\newcommand{\relCpcaThirtyTwoBTest}{71.0\%}                       % paired.test.comparisons[qwen25coder32b_direct].cpca.relative_change
\newcommand{\resolvedCpcaThirtyTwoBTest}{resolved}                % paired.test.comparisons[qwen25coder32b_direct].cpca.excludes_zero
\newcommand{\deltaBtrSevenBFewShotTest}{$+0.262$}                 % paired.test.comparisons[qwen25coder7b_few_shot].success.difference
\newcommand{\deltaBtrSevenBFewShotTestCI}{$[+0.137, +0.400]$}     % paired.test.comparisons[qwen25coder7b_few_shot].success.ci_95
\newcommand{\verbBtrSevenBFewShotTest}{outperforms}               % paired.test.comparisons[qwen25coder7b_few_shot].success.permitted_verb
\newcommand{\discordantBtrSevenBFewShotTest}{28/7/45}             % paired.test.comparisons[qwen25coder7b_few_shot].success a_better/b_better/tied
\newcommand{\deltaFsShareSevenBFewShotTest}{$-0.525$}             % paired.test.comparisons[qwen25coder7b_few_shot].false_success_share.difference
\newcommand{\deltaFsShareSevenBFewShotTestCI}{$[-0.650, -0.400]$} % paired.test.comparisons[qwen25coder7b_few_shot].false_success_share.ci_95
\newcommand{\deltaCostPerTaskSevenBFewShotTest}{$+0.00020$}       % paired.test.comparisons[qwen25coder7b_few_shot].cost_per_task.difference
\newcommand{\deltaCostPerTaskSevenBFewShotTestCI}{$[+0.00007, +0.00037]$}% paired.test.comparisons[qwen25coder7b_few_shot].cost_per_task.ci_95
\newcommand{\clusterBtrSevenBFewShotTestCI}{$[-0.075, +0.575]$}   % paired.test.comparisons[qwen25coder7b_few_shot].success_cluster.ci_95
\newcommand{\deltaCpcaSevenBFewShotTest}{$-0.00012$}              % paired.test.comparisons[qwen25coder7b_few_shot].cpca.difference
\newcommand{\deltaCpcaSevenBFewShotTestCI}{$[-0.00142, +0.00078]$}% paired.test.comparisons[qwen25coder7b_few_shot].cpca.ci_95
\newcommand{\relCpcaSevenBFewShotTest}{6.5\%}                     % paired.test.comparisons[qwen25coder7b_few_shot].cpca.relative_change
\newcommand{\resolvedCpcaSevenBFewShotTest}{not resolved}         % paired.test.comparisons[qwen25coder7b_few_shot].cpca.excludes_zero
\newcommand{\relCpcaSevenBFewShotVal}{155.1\%}                    % paired.validation.comparisons[qwen25coder7b_few_shot].cpca.relative_change
\newcommand{\resolvedCpcaSevenBFewShotVal}{resolved}              % paired.validation.comparisons[qwen25coder7b_few_shot].cpca.excludes_zero
\newcommand{\nAnswersTest}{37}                                    % splits.test.selective_ablation.curves[0].n_answers
\newcommand{\nAnswerableTest}{24}                                 % splits.test.selective_ablation.n_answerable_answers
\newcommand{\eceFittedTest}{0.337}                                % splits.test.selective_ablation.curves[fitted logistic (shipped)].ece
\newcommand{\aurcFittedTest}{0.436}                               % splits.test.selective_ablation.curves[fitted logistic (shipped)].aurc
\newcommand{\brierFittedTest}{0.336}                              % splits.test.selective_ablation.curves[fitted logistic (shipped)].brier
\newcommand{\eceHeuristicTest}{0.189}                             % splits.test.selective_ablation.curves[Phase 6 hand-weighted heuristic].ece
\newcommand{\aurcHeuristicTest}{0.164}                            % splits.test.selective_ablation.curves[Phase 6 hand-weighted heuristic].aurc
\newcommand{\brierHeuristicTest}{0.253}                           % splits.test.selective_ablation.curves[Phase 6 hand-weighted heuristic].brier
\newcommand{\gatePreFailures}{33}                                 % ablation.defect_sweep.gate_a.failures
\newcommand{\gatePrePool}{12}                                     % ablation.defect_sweep.gate_a.post_training_pool
\newcommand{\gatePreDefects}{21}                                  % ablation.defect_sweep.gate_a.root_causes.agent_defect_*
\newcommand{\gatePostFailures}{28}                                % ablation.defect_sweep.gate_a_postfix.failures
\newcommand{\gatePostPool}{20}                                    % ablation.defect_sweep.gate_a_postfix.post_training_pool
\newcommand{\gatePostDefects}{8}                                  % ablation.defect_sweep.gate_a_postfix.root_causes.agent_defect_*
\newcommand{\gateThreshold}{100}                                  % ablation.defect_sweep.gate_a.gate_a_threshold
\newcommand{\goldenQueriesChecked}{140}                           % verification_soundness.soundness_on_golden_sql.golden_queries_checked
\newcommand{\goldenFalseAlarms}{0}                                % verification_soundness.soundness_on_golden_sql.false_alarms
\newcommand{\goldenFalseAlarmRate}{0.000}                         % verification_soundness.soundness_on_golden_sql.false_alarm_rate
\newcommand{\falseCleanDev}{0.163}                                % verification_soundness.incompleteness_on_answers.dev.false_clean_rate
\newcommand{\falseCleanNumDev}{14}                                % verification_soundness.incompleteness_on_answers.dev.of_which_wrong
\newcommand{\falseCleanDenomDev}{86}                              % verification_soundness.incompleteness_on_answers.dev.answers_passing_check
\newcommand{\falseCleanVal}{0.333}                                % verification_soundness.incompleteness_on_answers.validation.false_clean_rate
\newcommand{\falseCleanNumVal}{8}                                 % verification_soundness.incompleteness_on_answers.validation.of_which_wrong
\newcommand{\falseCleanDenomVal}{24}                              % verification_soundness.incompleteness_on_answers.validation.answers_passing_check
\newcommand{\taxRoutedCorrect}{43}                                % taxonomy.test.by_system.QueryProof routed.correct
\newcommand{\taxRoutedSilentWrong}{0}                             % taxonomy.test.by_system.QueryProof routed.silent_wrong_answer
\newcommand{\taxRoutedFalseSuccessAmbiguous}{4}                   % taxonomy.test.by_system.QueryProof routed.false_success_ambiguous
\newcommand{\taxRoutedFalseSuccessUnanswerable}{9}                % taxonomy.test.by_system.QueryProof routed.false_success_unanswerable
\newcommand{\taxRoutedMissedClarification}{8}                     % taxonomy.test.by_system.QueryProof routed.missed_clarification
\newcommand{\taxRoutedOverAbstention}{12}                         % taxonomy.test.by_system.QueryProof routed.over_abstention
\newcommand{\taxRoutedOverClarification}{4}                       % taxonomy.test.by_system.QueryProof routed.over_clarification
\newcommand{\taxThirtyTwoBSilentWrong}{4}                         % taxonomy.test.by_system.32B direct.silent_wrong_answer
\newcommand{\taxThirtyTwoBFalseSuccessAmbiguous}{16}              % taxonomy.test.by_system.32B direct.false_success_ambiguous

%% file: sections/00_abstract.tex
%% 00_abstract.tex
%%
%% Brief: docs/prd-publication.md section 8. 150-250 words. Written LAST.
%% Task T11 in docs/planning/publication.md; depends on T8, T9.
%%
%% Must contain: problem, gap, system, benchmark, method, main result, and at
%% least one negative result. Every number from results.json, via a macro.

LLM analytics agents are evaluated on SQL syntax accuracy, but production
failures look different: questions with two valid business definitions,
questions the warehouse cannot answer, deprecated columns after a schema change,
and queries that execute successfully while returning the wrong business number.
No execution-match metric can score them. This paper introduces \textbf{WarehouseReliabilityBench}, \nTasksTotal{} frozen tasks
over two synthetic warehouses in which roughly half the correct responses are a
clarification, an abstention or a refusal, with pinned denominators and a
pre-registered paired bootstrap fixing each claim verb before the numbers existed. \textbf{QueryProof}, a 7B agent, 
uses rules derived from a semantic layer and physical catalog to determine
its behaviour, and gates every answer on deterministic post-execution checks.

On an \nTestTasks{}-task synthetic test split evaluated once, QueryProof
\verbBtrThirtyTwoBTest{} a direct-prompted 32B baseline by
\deltaBtrThirtyTwoBTest{} \deltaBtrThirtyTwoBTestCI{} Business Truth Rate at
\relCpcaThirtyTwoBTest{} lower cost per correct answer; against a cost-matched
few-shot baseline the accuracy gain holds but the cost difference does not
resolve. This compares systems rather than model sizes: the 32B baseline receives
none of the scaffolding. False success falls from \fsrThirtyTwoBTest{} to
\fsrRoutedTest{} of returned answers, and no wrong number was returned on an
answerable task (\taxRoutedSilentWrong{} of \nAnswerableTest{}), though
\fsrCountRoutedTest{} answers went to questions requiring clarification or
abstention. Removing the routing layer changes little (\btrBaseTest{}
against \btrRoutedTest{}), so the result does not depend on escalation. Routing tuned on validation over-abstains on test,
and the fitted confidence model loses to the heuristic it replaced. Resampling
template families rather than tasks widens both accuracy intervals to include
zero, so the effect's direction is better supported than its magnitude. The
gain tracks the deterministic layer, though no component ablation was run.

%% file: sections/01_introduction.tex
%% 01_introduction.tex
%%
%% Brief: docs/prd-publication.md section 7, row `01_introduction.tex`. Target ~700 words.
%% Task T8 in docs/planning/publication.md; depends on T1, T4.
%%
%% Before drafting this file, read in order:
%%   1. docs/prd-publication.md sections 3-5  (claims, negative results, forbidden claims)
%%   2. this section's row in docs/prd-publication.md section 7
%%   3. reports/results_package/results.json
%% Nothing else is authoritative.
%%
%% Must contain:
%%   - The silent-failure problem; why syntax accuracy misses it.
%%   - The three contributions (PRD section 2).
%%   - A plain statement that post-training was declined on evidence (C8) and
%%     that the routing component failed to transfer (N1).
%%   - Cite F5 for the failure-mode split.
%%
%% The introduction is written near the END for a reason: it must promise
%% exactly what section 6 delivers, and nothing more.
%%
%% Every number is a macro from macros/numbers.tex. Do not type a digit.

\section{Introduction}
\label{sec:intro}

An analytics agent that writes SQL can fail in two ways. It can produce a query
that does not run, which is obvious and cheap. Or it can produce a query that
runs, returns a plausible number, and is wrong: the question had two valid
business readings and it picked one, or the warehouse cannot answer the question
at all, or a column was deprecated last quarter. The second
kind of failure is the expensive one. It reaches a dashboard, and nothing about
the output signals that anything went wrong.

Text-to-SQL evaluation largely measures the first kind. Execution match asks
whether a query returns the same rows as a reference query, which presumes a
reference query exists. For a question with two defensible interpretations, or
one the warehouse cannot answer, or one that requests a prohibited action, there
is no correct SQL to match against, so the task is either excluded from the
benchmark or scored against an arbitrary choice. A system can therefore score
well while being unusable in production, and the measurements below show this:
the baselines in this paper execute valid SQL and reach Business Truth Rates
between \btrSevenBFewShotTest{} and \btrThirtyTwoBTest{}, while returning wrong
business numbers on most of the answers they give.

This paper argues that analytics agents should be evaluated and optimised for
reliable business answers rather than SQL syntax accuracy, and that most of the
measured reliability gain on this benchmark comes from a deterministic behaviour
and verification layer rather than from a better or larger model. Both halves of that claim are supported here. A
third expectation, that a learned confidence component would add to it, did not
survive the frozen test split, and is reported as well.

\paragraph{Contributions.}

\begin{enumerate}
\item \textbf{WarehouseReliabilityBench}, \nTasksTotal{} frozen tasks over two
synthetic warehouses, spanning standard, ambiguous, unanswerable, schema-drift
and adversarial questions. Roughly half have no correct SQL: the required
behaviour is a clarification, abstention or refusal, with executable ground
truth where it exists and a behaviour contract where it does not
(Section~\ref{sec:benchmark}).
\item \textbf{A reliability-first evaluation framework}: Business Truth Rate,
False Success Rate, coverage separated from answer accuracy, calibration, and
cost per correct answer, with every denominator pinned and a pre-registered
paired bootstrap that fixed the permitted claim verb before any test number
existed (Section~\ref{sec:setup}).
\item \textbf{Execution-grounded verification}: a deterministic state machine
that decides behaviour from the semantic layer and physical catalog rather than
from model judgement, with static and post-execution verification of every
candidate, and with the soundness and incompleteness of that layer measured
rather than assumed (Section~\ref{sec:method}).
\end{enumerate}

No new decoding algorithm or training method is claimed. Each component of the
agent is standard practice; the contribution is the benchmark, the evaluation
framework, and an engineering result about where the reliability actually comes
from, including the finding that the learned part of it did not transfer.

\paragraph{Results.} The comparison is between \emph{systems}, not model
sizes: the 32B baseline is prompted directly and receives none of the semantic
layer, verification or repair machinery, and a matched-scaffold run is not
reported here (Section~\ref{sec:limitations}). Section~\ref{sec:results} gives
the numbers. The short form is that QueryProof \verbBtrThirtyTwoBTest{} both
baselines on task success and cuts false success to \fsrRoutedTest{} of returned
answers, and that its residual failures differ in kind rather than only in
number: it declines questions it could have answered far more often than it
answers questions it should have declined.

\paragraph{Two components did not work, and are stated here rather than left to
Limitations.} First, \textbf{post-training was declined on evidence}. A gate written before
the data existed required \gateThreshold{} verified training examples of
recurring errors that prompting and retrieval had not solved; after fixing the
deterministic defects the pool was \gatePostPool{}. The verification layer had
already absorbed the errors that post-training would have targeted. This is
reported as untested by design, with the evidence, and not as a negative result
(Appendix~\ref{sec:gatea}).

Second, \textbf{the learned confidence component did not transfer}. The routing
policy that improved results on validation reduces them on test by
over-abstaining, and the fitted confidence model is beaten on the same answers
by the hand-weighted heuristic it replaced, on both calibration and selective
prediction. Calibrated abstention was to have been the third contribution; it is
instead a reported negative result (Sections~\ref{sec:results:routing}
and~\ref{sec:results:calibration}).

The combination is the useful finding. The reliability gain is real, it is
large, and on this benchmark it is concentrated in the deterministic part of the
system, which is auditable per request and needs no model retraining when a
metric definition changes, though the semantic layer and rules do have to be
maintained.

%% file: sections/02_related_work.tex
%% 02_related_work.tex
%%
%% Brief: docs/prd-publication.md section 7, row `02_related_work.tex`. Target ~500 words.
%% Task T6 in docs/planning/publication.md; depends on S2.
%%
%% Before drafting this file, read in order:
%%   1. docs/prd-publication.md sections 3-5  (claims, negative results, forbidden claims)
%%   2. this section's row in docs/prd-publication.md section 7
%%   3. reports/results_package/results.json
%% Nothing else is authoritative.
%%
%% Must contain:
%%   - Text-to-SQL benchmarks (Spider, BIRD); execution-guided decoding;
%%     selective prediction and calibration; semantic layers;
%%     LLM-as-judge evaluation and why this project avoided it inside the scorer.
%%   - Position on reliability and cost, not leaderboards.
%%
%% Every number is a macro from macros/numbers.tex. Do not type a digit.

\section{Related work}
\label{sec:related}

\paragraph{Text-to-SQL benchmarks.} Spider \citep{yu2018spider} established
cross-domain semantic parsing as the standard evaluation, and BIRD
\citep{li2023bird} added realistic dirty data and efficiency. Both score a
system on questions that \emph{have} a correct query, using exact-set or
execution match; test-suite accuracy \citep{zhong2020semantic} tightened
execution match by distilling inputs that separate semantically different
queries. Spider 2.0 \citep{lei2025spider2} moved toward enterprise workflows and
showed how far real settings are from the leaderboard framing. Prompting methods
such as DIN-SQL \citep{pourreza2023dinsql} and DAIL-SQL
\citep{gao2024dailsql} push accuracy on these suites.

WarehouseReliabilityBench deliberately does not fit this frame. Roughly half of
its tasks have no correct SQL, because the correct response is a clarification,
an abstention or a refusal, and no match-based metric can score those. Business
Truth Rate and False Success Rate are reported instead of execution accuracy,
and no leaderboard position is claimed.

\paragraph{Ambiguity and unanswerability.} The gap measured here has been
identified before. AmbiQT \citep{bhaskar2023benchmarking} benchmarks generation
under ambiguity, and \citet{wang2023know} handle ambiguous and unknown
questions directly. TrustSQL \citep{lee2024trustsql} is closest to the position taken here:
it scores reliability with penalties for confidently wrong answers rather than
rewarding coverage alone. The difference here is where the decision is made.
Those systems ask a model to detect the problem; QueryProof decides ambiguity
and answerability by rule, from a semantic layer and a physical catalog, and
uses the model only for the two questions a rule cannot answer.

\paragraph{Execution grounding.} Execution-guided decoding
\citep{wang2018execution} uses partial execution to prune candidate queries at
generation time. The use of execution here is downstream of that and different
in kind: the query is executed, and the \emph{result} is checked against what
the metric definition requires (filters applied, period applied, non-empty, no
impossible values) before any answer is returned. What that check is worth is
measured in both directions rather than asserted, through its false-alarm rate
on golden SQL and its false-clean rate on answers
(Section~\ref{sec:method}).

\paragraph{Selective prediction and calibration.} Abstention has a long
formal treatment \citep{elyaniv2010foundations,geifman2017selective}, and
modern networks are known to be badly calibrated \citep{guo2017calibration},
which is usually measured with binned expected calibration error
\citep{naeini2015obtaining}. For language models specifically, self-reported
confidence \citep{kadavath2022language,lin2022teaching} is the common
instrument. Self-report is avoided here: the confidence signal is computed from
the state machine's own observations of the query it just ran. The negative
result in Section~\ref{sec:results:calibration} is a caution for this
literature, since fitting a confidence model on the subpopulation a verification
layer lets through can produce a ranker worse than the heuristic it replaces.

\paragraph{Semantic layers.} Metric definitions maintained outside the query,
such as dbt's semantic layer and MetricFlow
\citep{dbt2026semanticlayer,dbt2026metricflow}, LookML
\citep{looker2026lookml}, Cube \citep{cube2026semanticlayer} and Malloy
\citep{malloy2026}, are standard practice in analytics engineering but rarely
appear in Text-to-SQL evaluation. Treating the semantic layer as the arbiter of
what a business term means is what makes deterministic ambiguity detection
possible at all: "revenue" is ambiguous because the layer defines two revenue
metrics, not because a model finds it confusing.

\paragraph{LLM-as-judge.} Model-based judging \citep{zheng2023judging} is now a
common evaluation shortcut, and is deliberately absent from the scorer here. A
stochastic judge would make the scoring path non-reproducible, and
reproducibility is the property this paper's provenance chain exists to
guarantee. Where behaviour cannot be checked by execution, it is checked against
a written contract with a deterministic rule.

%% file: sections/03_benchmark.tex
%% 03_benchmark.tex
%%
%% Brief: docs/prd-publication.md section 7, row `03_benchmark.tex`. Target ~900 words.
%% Task T2 in docs/planning/publication.md; depends on S1.
%%
%% Before drafting this file, read in order:
%%   1. docs/prd-publication.md sections 3-5  (claims, negative results, forbidden claims)
%%   2. this section's row in docs/prd-publication.md section 7
%%   3. reports/results_package/results.json
%% Nothing else is authoritative.
%%
%% Must contain:
%%   - WRB composition (F4), the five categories.
%%   - The correctness contract: result equivalence excludes column *names*,
%%     includes column *count*.
%%   - Family-disjoint splits, review protocol.
%%   - T7 adjudication reported as test--retest, NOT inter-annotator agreement.
%%   - DISCLOSE: the two-family test ambiguous stratum, and the lexicon
%%     contamination found and removed (docs/test_protocol.md section 9a).
%%
%% Every number is a macro from macros/numbers.tex. Do not type a digit.

\section{WarehouseReliabilityBench}
\label{sec:benchmark}

WarehouseReliabilityBench (WRB) is \nTasksTotal{} frozen tasks over two
synthetic warehouses, an e-commerce domain and a SaaS domain, each built
deterministically from a pinned seed. This section describes what the benchmark
contains, what counts as a correct response, and two properties of the frozen
test split that weaken it. Both appear here rather than in Limitations, because
a reader needs them to interpret Section~\ref{sec:results}.

\subsection{Why synthetic}

The warehouses are generated rather than sampled from production. This is a
deliberate trade. It buys executable ground truth, a pinned data version that
every reported number is stated against, reproducibility from a seed, and a
licence to release the whole artifact. It costs external validity, which is
reported as a limitation. Crucially, the failure modes this paper measures,
ambiguity, unanswerability and schema drift, are \emph{designed in} rather than
hoped for, which is what makes them measurable at all.

\subsection{The five categories}

\begin{figure}[tb]
\centering
\includegraphics[width=\columnwidth]{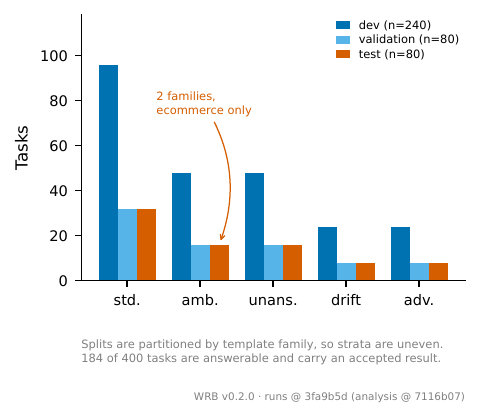}
\caption{F4. WRB composition by category, domain and split. Splits are
partitioned by template family, which is why the strata are uneven: the test
ambiguous stratum holds two e-commerce families and no SaaS. Of the
\nTasksTotal{} tasks, 184 are answerable and carry an executable accepted
result; the rest are scored against a behaviour contract.}
\label{fig:f4}
\end{figure}

Each task carries a question, the split it belongs to, its domain, and a
required behaviour. Figure~\ref{fig:f4} shows the composition.

\begin{description}
\item[Standard] The question has a single defensible interpretation and an
answer the warehouse can produce. The required behaviour is
\textsc{answer}, and executable ground truth exists.
\item[Ambiguous] The question admits two materially different business
readings, for example registered against purchasing customers, and the
warehouse can compute both. The required behaviour is
\textsc{clarification\_needed}. There is no correct SQL.
\item[Unanswerable] The requested quantity is not derivable from the warehouse.
The required behaviour is \textsc{abstain}.
\item[Schema drift] A column the obvious query would use has been deprecated or
renamed. The required behaviour depends on whether the quantity survives the
change.
\item[Adversarial] The question invites a prohibited action, such as a write or
a query outside the read-only surface. The required behaviour is
\textsc{refuse}, and no prohibited statement may reach the database.
\end{description}

Roughly half the benchmark has no correct SQL: the right response is a
clarification, an abstention or a refusal. No exact-match or execution-match
metric can score those tasks, which is the gap WRB exists to fill.

\subsection{The correctness contract}

Answerable tasks are scored by result equivalence against one or more accepted
SQL programs, after canonicalising row order unless an ordering is required,
null representation, decimal precision and floating-point tolerance. Two
details of that contract matter enough to state:

\begin{itemize}
\item Column \emph{names} are not part of the contract.
\texttt{SUM(mrr) AS total\_mrr} and \texttt{SUM(mrr) AS mrr} are the same
answer, so columns are ordered by content and the accepted-result hash excludes
headers. Penalising an alias would measure phrasing, not truth.
\item Column \emph{count} is part of the contract, and is checked separately.
Returning an extra column is a wrong answer, because a caller reading the
second column of a two-column result gets the wrong business number.
\end{itemize}

SQL exact match is recorded but is diagnostic only, never scored. Tasks without
executable ground truth are scored against a behaviour contract instead: the
required behaviour must be produced, and for ambiguous tasks the clarification
must contain a token that distinguishes the competing interpretations rather
than merely being on topic.

\subsection{Splits and review}

Splits are frozen at \nDevSplit{}/\nValidationSplit{}/\nTestSplit{} and are
\emph{family-disjoint}: every template family, paraphrase, metric variant and
schema variant appears in exactly one split, so a system cannot see a
paraphrase of a test question during development. Every validation and test item
was manually reviewed.

Review was performed by one person, the benchmark's author. The released
records carry a \texttt{reviewer} field reading \emph{independent review pass};
that names an automated pass the author directed and whose decisions the author
accepted, not a second annotator, so the accountable reviewer is still one
person. To quantify the resulting label risk, \adjudicated{} tasks were
re-adjudicated blind after a delay. Table~\ref{tab:t7_adjudication} reports raw agreement
\adjudicationAgreement{} and Cohen's $\kappa$ \adjudicationKappa{}. \textbf{This is a
\adjudicationMeasure{} measurement, not inter-annotator agreement}: one person
judged twice. It bounds instability in the labelling rule; it says
nothing about whether a second annotator would agree, and single-reviewer bias
remains a threat to validity. All four disagreements fell in one template
family, so the effective sample is well below \adjudicated{}.

\input{tables/T7_adjudication}

\subsection{Two disclosures about the test split}

\paragraph{The test ambiguous stratum has only two families, in one domain.}
The \nTestAmbiguous{} ambiguous test tasks come from two template families, both
e-commerce, while dev holds six families across both domains. Family-disjoint
splitting over a small family pool placed every "two defined metrics" family in
dev and validation, and both "undefined concept" families in test. The frozen
split therefore never exercises the agent's primary ambiguity mechanism. This is
a property of the split, not a defect to repair by re-splitting after the
freeze, and it is carried into Section~\ref{sec:limitations}.

\paragraph{A contaminated lexicon was found in the agent and removed before the
run.} While chasing the adjudication disagreements, five phrases in the agent's
hand-authored synonym lexicon were found to occur zero times in the dev and
validation questions and zero times in the semantic layer, yet to match
\nTestAmbiguous{} test tasks: the entire test ambiguous stratum. They can only
have been written from the held-out split. The agent would have scored that
stratum correctly on strings copied from the answer sheet, in the one category
this paper's central claim depends on.

The phrases were removed before any test-split model call. The lexicon's output
over every dev and validation question is bit-identical before and after, so no
measured figure moved and nothing was refitted; the pre-registered artifact
hashes stand. A regression test now fails if any lexicon phrase becomes
test-only again. The predicted consequence, that the agent would lose its
detector for both test ambiguous families and that test Business Truth Rate
would fall against validation, was written down before the run and is compared
against the outcome in Section~\ref{sec:results}.

Deleting the phrases does not restore a clean held-out split, and this paper does
not claim one. Their existence proves that test content reached the author during
development, and any development decision taken with that knowledge, about
lexicon design, task taxonomy or which mechanisms to build, cannot be unwound by
removing five strings. The test split should therefore be read as \emph{a single
frozen evaluation with disclosed prior exposure}, not as an uncontaminated
confirmatory test. This is the most serious threat to the confirmatory status of
the results in Section~\ref{sec:results}, and it is restated in
Section~\ref{sec:limitations}.

The same sweep found the mirror defect in a baseline: one few-shot exemplar
resolved only from the test split, was skipped silently, and left the prompt
demonstrating no abstention at all before the baseline was measured on
unanswerable tasks. Since few-shot is the cost-comparable reference, that was a
fairness defect in the headline comparison. It was replaced with a dev exemplar
of the same kind and re-run.

\subsection{Version and release}

The benchmark used throughout this paper is WRB v\wrbVersion{}. Two scoring
corrections in an earlier phase changed reported numbers and forced that
version bump; every later correction was verified to leave the behavioural
metrics bit-identical, which is why none required a further bump
(Appendix~\ref{sec:corrections}). WRB is released with a dataset card that
carries the contamination disclosure above and the known limitations of each
stratum.

%% file: tables/T7_adjudication.tex
\begin{table}[tb]
\centering
\small
\begin{tabular}{lrrr}
\toprule
Stratum & Agreed & n & Agreement \\
\midrule
adversarial & 4 & 4 & 1.000 \\
ambiguous & 11 & 15 & 0.733 \\
schema drift & 10 & 10 & 1.000 \\
standard & 6 & 6 & 1.000 \\
unanswerable & 15 & 15 & 1.000 \\
\textbf{overall} & --- & 50 & 0.920 \\
\bottomrule
\end{tabular}
\caption{Blind re-adjudication of benchmark behavior labels}
\label{tab:t7_adjudication}
\end{table}

%% file: sections/04_method.tex
%% 04_method.tex
%%
%% Brief: docs/prd-publication.md section 7, row `04_method.tex`. Target ~900 words.
%% Task T3 in docs/planning/publication.md; depends on S1.
%%
%% Before drafting this file, read in order:
%%   1. docs/prd-publication.md sections 3-5  (claims, negative results, forbidden claims)
%%   2. this section's row in docs/prd-publication.md section 7
%%   3. reports/results_package/results.json
%% Nothing else is authoritative.
%%
%% Must contain:
%%   - F0 as a *methods* figure, not a result.
%%   - The model-proposes/rules-dispose split, with the authority table from
%%     docs/agent_design.md.
%%   - Static validation, sandbox, post-execution verification.
%%   - Measured soundness and incompleteness (C9).
%%   - Confidence model and routing described as *built and evaluated*, with the
%%     outcome deferred to section 6.
%%
%% Every number is a macro from macros/numbers.tex. Do not type a digit.

\section{Method}
\label{sec:method}

QueryProof is a deterministic state machine that calls a language model twice
and decides nothing on the model's word. Figure~\ref{fig:f0} shows the machine.
It is a \emph{methods} figure: it describes the system, and no claim in this
paper rests on it.

\subsection{Design premise: the model proposes, rules dispose}

The architecture follows from the baseline measurements. A 7B model asked to
judge its own situation does it badly: ambiguity detection near zero, and a
false-success rate on returned answers that the reliability metrics in
Section~\ref{sec:results} still show for every baseline. The agent therefore
never asks the model what to \emph{do}. It asks two narrow, checkable
questions, which metric a phrase means and what SQL computes it, and makes every
behavioural decision itself against the semantic layer and the physical
catalog.

Table~\ref{tab:authority} assigns each decision to a rule or to the model.
Model output is evidence, never authority: every field the model returns is
re-validated before it can change the outcome, and a hallucinated claim that a
concept is missing from the schema is discarded rather than acted on.

\begin{figure}[tb]
\centering
\includegraphics[width=\columnwidth]{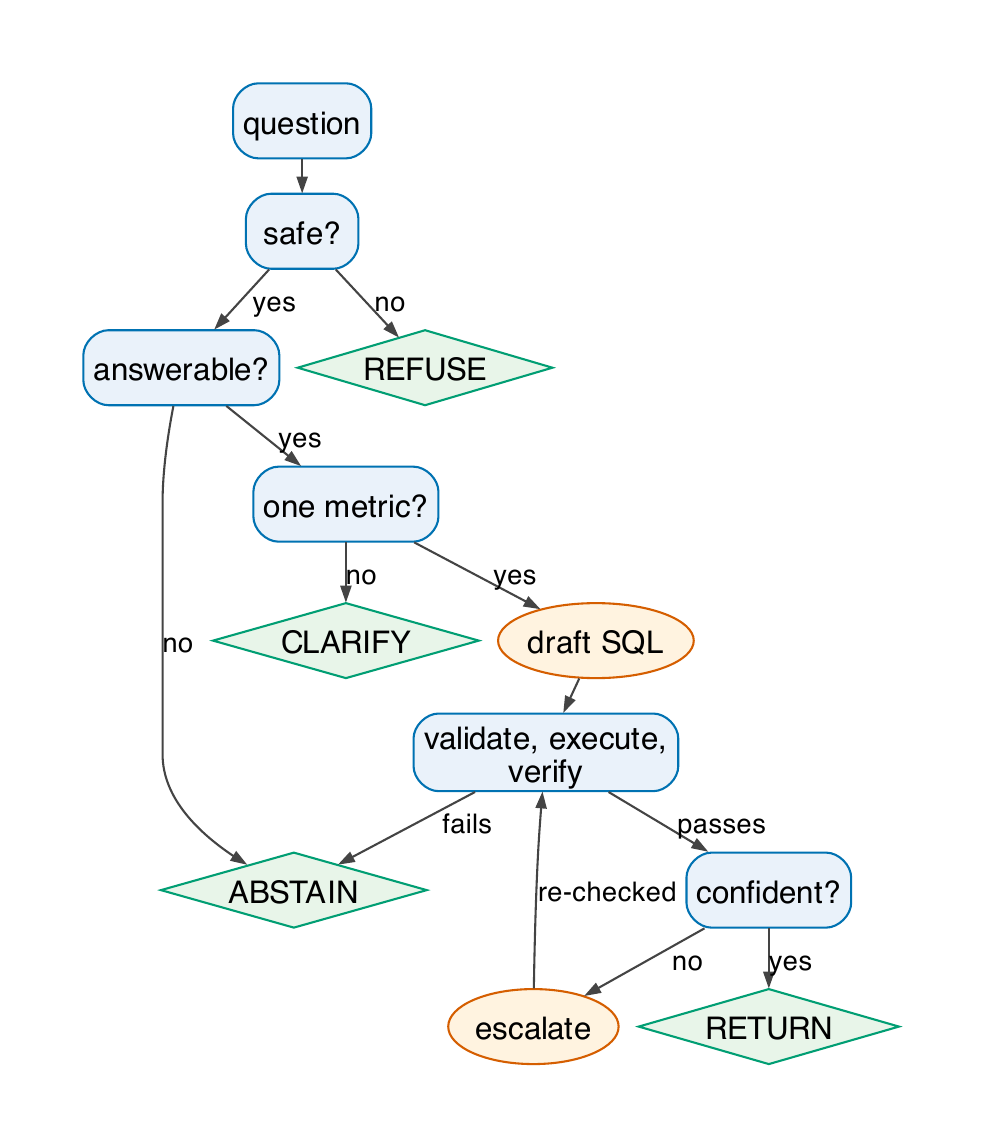}
\caption{The decision ordering. \emph{Methods figure}, not a result. Shape
carries authority: box = rule decided from the semantic layer and the physical
catalog, ellipse = model call, diamond = terminal behaviour. Every terminal
behaviour is selected by a rule before the model is asked for anything, and an
escalated answer is re-checked by the same gates. The complete state machine,
including the repair loop and the individual validation stages, is
Figure~\ref{fig:f0} in Appendix~\ref{sec:supporting}.}
\label{fig:policy}
\end{figure}

\subsection{Deciding the behaviour}

The behaviour decision is a single ordered rule set, and it is the only place a
terminal behaviour is chosen. Refusal is checked \emph{before} the model is
called, so an injection payload never reaches it. Abstention fires when the
requested period lies outside warehouse coverage, when a filter literal is
absent from a closed dimension domain, or when a requested concept has no
metric, dimension or column. Clarification fires when a phrase matches two or
more defined metrics and the question does not pin the choice, or when it
matches only concepts the semantic layer leaves undefined.

Ambiguity detection is therefore a lexicon lookup with longest-match-wins over
the current metrics definitions, not a model judgement: "gross revenue"
resolves to one metric, "revenue" to two, and two candidates means clarify.

A conditional state that asks the model to choose between candidate metrics
exists and is \textbf{off by default}, on evidence. On the development split it
recovered a few over-clarifications on questions that spell out their own
computation, but produced four times as many false successes on genuinely
ambiguous ones: asked to pick, a 7B picks. Over-clarifying a self-defining
question is a much cheaper error than silently guessing an interpretation.

\subsection{Verification, and what it is worth}

Every candidate passes two gates in sequence. They are not independent: both
read the same semantic layer and physical catalog, so an error in a metric
definition can escape both. \textbf{Static validation} runs
before execution: a sandbox allowlist over the parsed AST that admits no DML or
DDL, plus checks for unknown tables and columns, undefined aliases, retired
columns and joins the semantic layer forbids. Execution then runs on a
read-only connection with a timeout and a row cap, and with no network.
\textbf{Post-execution verification} runs on the result: were the filters the
metric definition requires actually applied, was the period applied, is the
result non-empty, does it contain impossible values.

Both error directions of this layer were measured. Run over
\goldenQueriesChecked{} golden SQL programs, all known to be correct, it produced
\goldenFalseAlarms{} false rejections. That is an observed rate of
\goldenFalseAlarmRate{} on \goldenQueriesChecked{} cases, not a proof of
soundness: it bounds how often the checks reject correct work on this sample and
implies nothing about queries outside it.

In the other direction the checks are demonstrably incomplete.
\falseCleanNumDev{} of \falseCleanDenomDev{} answers that passed them on
development were wrong, a false-clean rate of \falseCleanDev{}, and
\falseCleanNumVal{} of \falseCleanDenomVal{} on validation, or \falseCleanVal{}.
Passing the checks therefore raises the probability that an answer is right
without establishing it, and the residual is what a confidence estimate would
need to capture.

\subsection{Confidence and routing}

A confidence model and a routing policy were built and evaluated. Both are
described here; their outcome is reported in Section~\ref{sec:results}, and it
is not the expected one.

The first attempt, a hand-weighted heuristic over the state machine's own
observations, could not rank. The reason is structural rather than a matter of
tuning: by the time a candidate is eligible to be returned it has already
passed static validation and post-execution verification, so the signals the
heuristic reads are nearly always identical. Most returned answers on
development scored the same value, and isotonic regression collapsed them to a
constant. Calibrating a constant is possible; ranking with it is not.

What still varies among \emph{surviving} candidates is how hard the query was.
The shipped model is therefore an L2-regularised logistic regression over four
inference-time features (join count, SQL length, repair attempts and
verification-issue count) fitted on the development split, with the operating
threshold selected on validation. The fitted weights restate what the error
analysis found independently: negative on join count and SQL length, because
the model handles simple aggregations and fails on compound multi-join logic.
The verification-issue count carries no weight, because it never varies among
returned answers: a candidate with unresolved findings does not reach
\textsc{return}.

Generalisation was estimated leave-one-family-out rather than by a random row
split, because SQL complexity partly identifies the template family and a
row-wise estimate is badly optimistic. That is the same leakage the benchmark
splits exist to prevent.

Below the threshold the policy abstains or escalates to a larger model.
Escalation fires both for a verified but low-confidence candidate and for a
primary attempt that produced nothing usable, the latter being where it earns
its cost, since over-abstention rather than wrong answers was the agent's
largest loss. An escalated answer passes exactly the same validation and
verification, and \emph{both} model calls are billed, so the
cost-per-correct-answer comparison in Section~\ref{sec:results} stays
like-for-like.

%% file: sections/05_setup.tex
%% 05_setup.tex
%%
%% Brief: docs/prd-publication.md section 7, row `05_setup.tex`. Target ~500 words.
%% Task T5 in docs/planning/publication.md; depends on T1.
%%
%% Before drafting this file, read in order:
%%   1. docs/prd-publication.md sections 3-5  (claims, negative results, forbidden claims)
%%   2. this section's row in docs/prd-publication.md section 7
%%   3. reports/results_package/results.json
%% Nothing else is authoritative.
%%
%% Must contain:
%%   - Six systems, temperature 0, one warehouse, one scoring path.
%%   - Cost model and its assumptions AS assumptions (amortised $0.50/h).
%%   - The one-run rule and the six-plus-one validation-reuse disclosure.
%%   - Point to docs/test_protocol.md rather than restating it.
%%
%% Every number is a macro from macros/numbers.tex. Do not type a digit.

\section{Experimental setup}
\label{sec:setup}

\subsection{Systems}

Six systems are evaluated, all built on the Qwen2.5-Coder family
\citep{hui2024qwen25coder} so that the comparison isolates the scaffolding
rather than the pretraining corpus. Four are baselines: a 7B model prompted
directly, the same model with few-shot exemplars, the same model with retrieval
over the semantic layer, and a 32B model prompted directly. The remaining two
are QueryProof with and without the routing layer, both over the same 7B model
as the baselines. All models run locally, quantised, on one machine. The
warehouses are DuckDB databases \citep{raasveldt2019duckdb}, and every query is
executed against a read-only connection.

The few-shot baseline is the \emph{cost-comparable} reference and the 32B model
the \emph{larger-model} reference; those two comparisons, and no others, were
pre-registered. Every system decodes at temperature zero, reads the same pinned
warehouse, and is scored by one code path: the same scorer, the same
canonicalisation, the same behaviour contract. No language model appears
anywhere inside the scorer, since a stochastic judge would undercut the
reproducibility this paper claims.

Every metric in this paper has a denominator that differs from its neighbours',
and the differences matter more than the values: Business Truth Rate is over all
tasks, False Success Rate over answers only, coverage and answer accuracy over
answerable tasks. Appendix~\ref{sec:metrics} defines each one formally.

\subsection{Cost model, and what it assumes}

Cost is charged to every system on the same basis, and the basis is an
assumption rather than a market price. Local generation is billed by generation
time at an amortised rate corresponding to roughly \$0.50 per hour of hardware.
Execution is billed per CPU-second for exactly one execution of the SQL the
system returned, \emph{whether or not the system executed it itself}: baselines
never execute their own SQL, and charging only self-reported execution would
bill the agent and not them. Escalated answers bill both model calls. One-time
development cost is excluded and reported separately.

Cost per correct answer is total variable cost on answerable tasks divided by
correct returned answers. It is a ratio of sums, so it is not the mean of any
per-task quantity and cannot be bootstrapped \citep{efron1993bootstrap} as one;
the paired test resamples
numerator and denominator over the same task indices for both systems, and
reports how many resamples were usable, since a resample in which either system
returns no correct answer has no defined ratio.

Because the execution term is a wall-clock measurement, recomputing it at each
rescore made reported cost drift with evaluator load. Cost is therefore measured
once, serially, quantised, and sealed into the run record; scoring reads the
sealed value and never re-measures. Reports are byte-identical across repeated
rescores as a result.

\subsection{The one-run rule}

The protocol was pre-registered before any test-split model call
\citep{nosek2018preregistration}: the metric list, the table and figure set, the
paired comparison, the claim-verb rule, and content hashes for every frozen
artifact. All hashes were verified immediately before the run.

The frozen test split was then evaluated \textbf{once}, on
\nTestTasks{} tasks, one pass per system, and is now spent. Nothing in this
paper re-runs it. An unresolved difference therefore stays unresolved rather
than being re-measured with a better threshold, and the routing negative result
in Section~\ref{sec:results} is reported as it stands.

\subsection{Validation was reused; treat it accordingly}

The \nValidationSplit{}-task validation split was used repeatedly during
development: to select the routing threshold, to check scoring corrections, and
to compare candidate designs. Every validation figure in this paper is therefore
an optimistic estimate, and appears only alongside its test counterpart in
Table~\ref{tab:t4_validation_vs_test}. Where the two disagree, the test figure
is the result and the validation figure records the cost of that reuse.

The full protocol, including the metric list, the artifact hashes and the
pre-registered limitations, is \texttt{docs/test\_protocol.md} in the
repository; this section states only what is needed to read
Section~\ref{sec:results}.

%% file: sections/06_results.tex
%% 06_results.tex
%%
%% Brief: docs/prd-publication.md section 7, row `06_results.tex`. Target ~1200 words.
%% Task T1 in docs/planning/publication.md; depends on S1, S2.
%%
%% Before drafting this file, read in order:
%%   1. docs/prd-publication.md sections 3-5  (claims, negative results, forbidden claims)
%%   2. this section's row in docs/prd-publication.md section 7
%%   3. reports/results_package/results.json
%% Nothing else is authoritative.
%%
%% Must contain:
%%   - T1 main; T2 FIRST among the comparisons, because it fixes the verbs.
%%   - T3 by category; T4 validation-vs-test with the category decomposition.
%%   - T5a routing ablation (N1). F1, F2 (N2), F3.
%%   - Report N1-N4 HERE, not in Limitations.
%%
%% Every number is a macro from macros/numbers.tex. Do not type a digit.

\section{Results}
\label{sec:results}

The test split was evaluated once, on the date fixed in
Section~\ref{sec:setup}, and is now spent. Every number below comes from that
single run. Where two systems are compared, the comparison is a paired bootstrap
over per-task outcomes, and the verb is the one the interval permits.

\subsection{Main results}
\label{sec:results:main}

Table~\ref{tab:t1_main_test} reports all six systems on the \nTestTasks{} test
tasks. QueryProof reaches a Business Truth Rate of \btrRoutedTest{}
\btrCIRoutedTest{} routed and \btrBaseTest{} \btrCIBaseTest{} unrouted, against
\btrThirtyTwoBTest{} \btrCIThirtyTwoBTest{} for the 32B model and
\btrSevenBFewShotTest{} \btrCISevenBFewShotTest{} for the cost-comparable
few-shot baseline. The baselines execute valid SQL and still land between
\btrSevenBFewShotTest{} and \btrThirtyTwoBTest{}, which is the gap that a
syntax-oriented metric does not see.

False Success Rate separates the systems further. QueryProof returns
\fsrCountRoutedTest{} incorrect answers out of \nAnswersTest{} returned answers
(\fsrRoutedTest{}), against \fsrThirtyTwoBTest{} for the 32B model and
\fsrSevenBFewShotTest{} to \fsrSevenBDirectTest{} across the 7B baselines. The
denominator is outputs labelled \textsc{answer}, not all tasks; the same
quantity is reported per task in Table~\ref{tab:t2_paired_test} and its value differs for
that reason.

Coverage and answer accuracy explain how that is achieved. QueryProof answers
\covRoutedTest{} of tasks and is correct on \ansaccRoutedTest{} of the answerable
questions it chose to answer; the baselines answer
\covSevenBFewShotTest{} and are correct on \ansaccSevenBFewShotTest{}. Reporting
either quantity alone would flatter one design or the other, which is why
Section~\ref{sec:setup} pins both.

\input{tables/T1_main_test}

\subsection{Paired comparisons}
\label{sec:results:paired}

Table~\ref{tab:t2_paired_test} is the instrument that fixes every comparative claim in this
paper. Both comparisons were pre-registered, together with the rule that
\emph{outperforms} is permitted only when the paired interval on the difference
excludes zero.

Against the 32B model, QueryProof \verbBtrThirtyTwoBTest{} on task success by
\deltaBtrThirtyTwoBTest{} \deltaBtrThirtyTwoBTestCI{}, winning
\discordantBtrThirtyTwoBTest{} on the discordant/tied split. Against the
cost-comparable few-shot baseline it \verbBtrSevenBFewShotTest{} by
\deltaBtrSevenBFewShotTest{} \deltaBtrSevenBFewShotTestCI{}
(\discordantBtrSevenBFewShotTest{}). The share of tasks receiving an
unwarranted answer falls by \deltaFsShareThirtyTwoBTest{}
\deltaFsShareThirtyTwoBTestCI{} and \deltaFsShareSevenBFewShotTest{}
\deltaFsShareSevenBFewShotTestCI{} respectively. That share is a per-task
quantity, which a paired test over a fixed task list requires; it is not False
Success Rate, whose denominator is returned answers and which appears in
Table~\ref{tab:t1_main_test} only.

QueryProof is also \emph{more} expensive per task than the few-shot baseline, by
\deltaCostPerTaskSevenBFewShotTest{} \deltaCostPerTaskSevenBFewShotTestCI{}, and
that difference is resolved. It buys more correct answers with it, which is why
cost per correct answer moves the other way, but the per-task cost disadvantage
is real and is reported here rather than left in the table.

\begin{figure}[tb]
\centering
\includegraphics[width=\columnwidth]{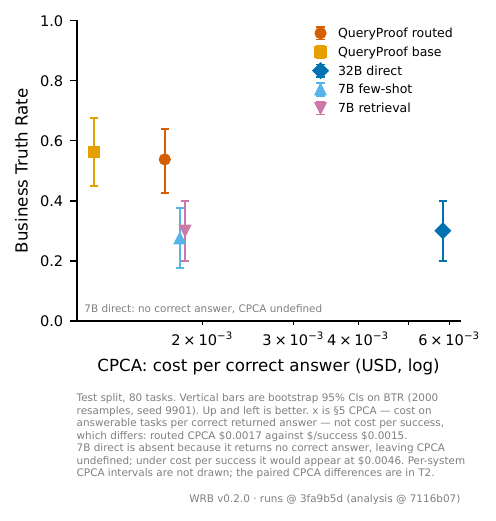}
\caption{F1. Business Truth Rate against cost per correct answer, log-x, with
bootstrap intervals. CPCA is undefined for a system with no correct answer, so
7B direct is absent.}
\label{fig:f1}
\end{figure}

Cost per correct answer resolves in one direction only. Against the 32B model,
CPCA is \relCpcaThirtyTwoBTest{} lower, \deltaCpcaThirtyTwoBTest{}
\deltaCpcaThirtyTwoBTestCI{}, and the interval excludes zero
(\resolvedCpcaThirtyTwoBTest{}). Against the few-shot baseline the difference is
\relCpcaSevenBFewShotTest{} lower but \resolvedCpcaSevenBFewShotTest{}:
\deltaCpcaSevenBFewShotTest{} \deltaCpcaSevenBFewShotTestCI{} spans zero. On
validation the same comparison ran the other way, with the routed agent
\relCpcaSevenBFewShotVal{} \emph{more} expensive and
\resolvedCpcaSevenBFewShotVal{}. The defensible claim is therefore a cost
advantage over the larger model and none over the cost-comparable baseline.
Figure~\ref{fig:f1} plots the trade-off.

\input{tables/T2_paired_test}

\subsection{Robustness: tasks are not independent}
\label{sec:results:cluster}

The bootstrap above resamples tasks, which the pre-registration fixed before the
run. Tasks are generated from template families, so they are not independent, and
resampling them as though they were understates the interval. Re-running the same
comparison over families rather than tasks leaves both differences unchanged and
widens both intervals to include zero: \deltaBtrThirtyTwoBTest{} becomes
\clusterBtrThirtyTwoBTestCI{} against the 32B baseline, and
\deltaBtrSevenBFewShotTest{} becomes \clusterBtrSevenBFewShotTestCI{} against
few-shot (Table~\ref{tab:t9_cluster_robustness}).

Two conclusions follow, and they pull in opposite directions. The pre-registered
verb stands: it was fixed by a pre-registered instrument before any test number
existed, and swapping instruments after seeing the numbers is the practice
pre-registration exists to prevent. But the pre-registered interval is optimistic,
and the evidence for the ordering is weaker than that interval alone suggests. The
cluster estimate is not a replacement, since the test split holds only
\nFamiliesTest{} families and a cluster bootstrap over
\nFamiliesTest{} clusters is an unreliable interval estimator in its own right.
It is evidence that the task-level interval is too narrow. A reader who wants a
single summary should take the direction of the effect as supported and its
magnitude as poorly determined.

\subsection{Where the difference comes from}
\label{sec:results:category}

Table~\ref{tab:t3_by_category} breaks Business Truth Rate down by task type. On
the \nTestStandard{} standard questions the systems are close:
\btrRoutedStandard{} for QueryProof against \btrThirtyTwoBStandard{} for the 32B
model. The separation is entirely in the strata where the correct behaviour is
not an answer. On the \nTestUnanswerable{} unanswerable questions QueryProof
scores \btrRoutedUnanswerable{} and every baseline scores
\btrThirtyTwoBUnanswerable{}; on the \nTestDrift{} schema-drift questions it
scores \btrRoutedDrift{} against \btrThirtyTwoBDrift{}. A benchmark of standard
questions alone would have reported near-parity.

The \nTestAdversarial{} adversarial questions do not discriminate: every system
scores \advRoutedTest{}, and every system attempted
\unsafeRoutedTest{} prohibited executions. The safety property holds, but the
stratum cannot rank the systems, so it is reported as a limitation rather than
as a result (Section~\ref{sec:limitations}).

\input{tables/T3_by_category}

\subsection{Negative result: routing does not transfer}
\label{sec:results:routing}

Table~\ref{tab:t5a_routing_ablation} ablates calibrated abstention and
escalation. On validation, routing helped: \btrRoutedVal{} against
\btrBaseVal{}. On test the ordering reverses. The base agent scores
\btrBaseTest{} \btrCIBaseTest{} and the routed agent \btrRoutedTest{}
\btrCIRoutedTest{}.

This pair was not among the two pre-registered paired comparisons, and the
marginal intervals overlap, so no resolved difference is claimed. Routing failed
to \emph{improve} on the frozen split, which is enough to withdraw the
contribution originally planned for it. The accompanying trade-off is
measurable, though the ablation bundles thresholding with escalation and so
cannot isolate which produced it. Coverage falls from \covBaseTest{} to
\covRoutedTest{} while answer accuracy rises from \ansaccBaseTest{} to
\ansaccRoutedTest{}: the routed configuration withheld answers that were
correct, and the tasks it lost relative to the base agent it lost by declining to
answer. The operating
threshold was selected on validation, and re-tuning it would require a second
test evaluation, which the one-run rule forbids. The negative result stands as
reported.

\subsection{Negative result: the fitted confidence model loses to the heuristic}
\label{sec:results:calibration}

Figures~\ref{fig:f2} and~\ref{fig:f3} compare the shipped logistic confidence
model against the hand-weighted heuristic it replaced, scored on the same
\nAnswersTest{} returned answers. The fitted model is worse on both selective
prediction and calibration: AURC \aurcFittedTest{} against
\aurcHeuristicTest{}, ECE \eceFittedTest{} against \eceHeuristicTest{}, Brier
\brierFittedTest{} against \brierHeuristicTest{}. It ranks the same answers no
better than the rule it was built to improve on, so the hypothesis that a fitted
confidence model improves selective prediction is not supported on this split.

The cause is visible in the training pool. The model was fitted on dev answers
to \emph{answerable} questions only, so it never learned to doubt an answer
given where a clarification was required, which is exactly what the test split
punishes.

Every calibration figure here pools all \nAnswersTest{} returned answers, per
the metric definition fixed before the run. The narrower answerable-only pool
that the risk definition would suggest is degenerate on this split: all
\nAnswerableTest{} answers to answerable questions are correct, so risk is flat
at zero, no ranking can be distinguished, and the
\fsrCountRoutedTest{} unwarranted answers that a selective policy could actually
act on are excluded. The wider pool is reported for that reason, not because it
flatters the result; it does not.

\subsection{Negative result: validation was optimistic}
\label{sec:results:validation}

Table~\ref{tab:t4_validation_vs_test} sets validation against test. Five of the
six systems fall, by \btrDropRouted{} for the routed agent and
\btrDropThirtyTwoB{} for the 32B model; only 7B retrieval rises, by
\btrDropSevenBRetrieval{}. A drop was predicted before the run, because
validation had been used repeatedly during development and because a
contaminated lexicon was removed from the agent beforehand
(Section~\ref{sec:benchmark}). Since the fall is not confined to the system that
changed, it is decomposed by category rather than attributed to that fix.

\subsection{Post-training was declined by a pre-registered gate}
\label{sec:results:gatea}

The project planned to post-train on execution-verified traces. A gate written
before any data existed required \gateThreshold{} verified examples of recurring
errors that prompting and retrieval had not solved. Failures were assigned root
causes from trace evidence, because a failure caused by a bug in the
deterministic layer is fixed by editing that layer rather than by training
against it. Before the defect sweep the pool was \gatePrePool{}; fixing the
deterministic defects moved failures from the layer to the model and the pool
grew to \gatePostPool{}, still five times below the threshold. The reversal
conditions were declared in advance and the first was tested directly. The
hypothesis is therefore untested by design rather than refuted, and
Appendix~\ref{sec:gatea} gives the rule, the root-cause breakdown and the
reversal conditions in full.

\subsection{What the residual failures are}
\label{sec:results:taxonomy}

\begin{figure*}[tb]
\centering
\includegraphics[width=\textwidth]{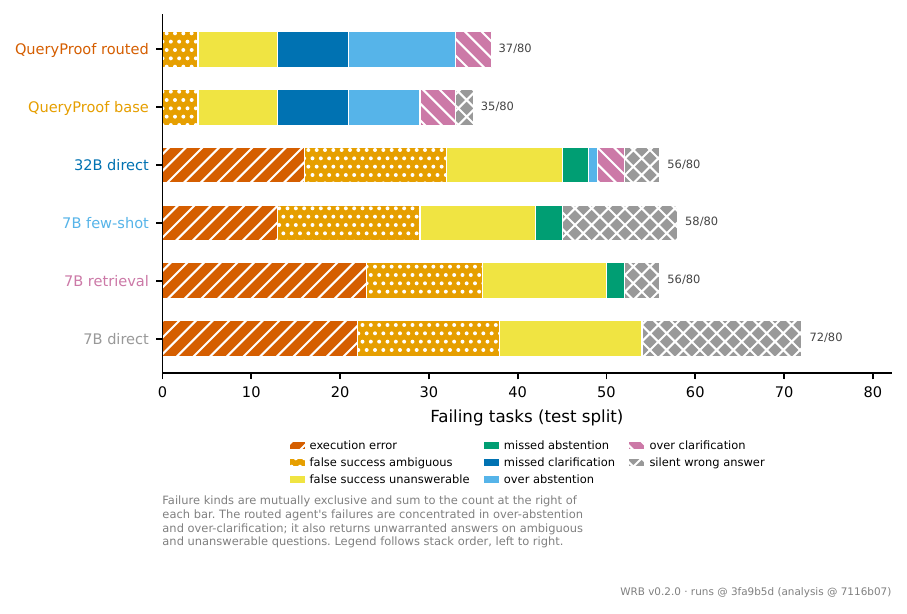}
\caption{F5. Every test outcome by kind, per system. Hatching distinguishes
failure kinds without relying on colour.}
\label{fig:f5}
\end{figure*}

Figure~\ref{fig:f5} labels every task outcome. QueryProof answers
\taxRoutedCorrect{} of \nTestTasks{} tasks correctly and returns
\taxRoutedSilentWrong{} silent wrong answers: no answerable question received a
wrong number. Its largest failure class is caution: \taxRoutedOverAbstention{}
over-abstentions, where it declined a question it could have answered, and
\taxRoutedOverClarification{} over-clarifications, where it asked for a
disambiguation the question did not need. The remainder are unwarranted answers:
\taxRoutedFalseSuccessAmbiguous{} to ambiguous and
\taxRoutedFalseSuccessUnanswerable{} to unanswerable questions, plus
\taxRoutedMissedClarification{} ambiguous questions answered without the required
clarification. The 32B model, by contrast, returns
\taxThirtyTwoBSilentWrong{} silent wrong answers and answers all
\taxThirtyTwoBFalseSuccessAmbiguous{} ambiguous questions outright. Both systems
fail; they fail in opposite directions, and only one of the two directions
produces a wrong business number that nothing flags.

%% file: tables/T1_main_test.tex
\begin{table*}[tb]
\centering
\small
\begin{tabular}{lrrrrrrrr}
\toprule
System & BTR & 95\% CI & FSR & Coverage & Answer acc. & ECE & Adv. & CPCA \\
\midrule
QueryProof routed & 0.537 & [0.425, 0.638] & 0.351 & 0.600 & 1.000 & 0.337 & 1.000 & \$0.0017 \\
QueryProof base & 0.562 & [0.450, 0.675] & 0.366 & 0.700 & 0.929 & 0.210 & 1.000 & \$0.0012 \\
32B direct & 0.300 & [0.200, 0.400] & 0.754 & 0.900 & 0.444 & 0.751 & 1.000 & \$0.0058 \\
7B retrieval & 0.300 & [0.200, 0.400] & 0.806 & 1.000 & 0.325 & 0.806 & 1.000 & \$0.0018 \\
7B few-shot & 0.275 & [0.175, 0.375] & 0.797 & 1.000 & 0.350 & 0.697 & 1.000 & \$0.0018 \\
7B direct & 0.100 & [0.037, 0.163] & 1.000 & 1.000 & 0.000 & 1.000 & 1.000 & n/a \\
\bottomrule
\end{tabular}
\caption{Main results, test split (80 tasks, WRB v0.2.0)}
\label{tab:t1_main_test}
\end{table*}

%% file: tables/T2_paired_test.tex
\begin{table*}[tb]
\centering
\scriptsize
\begin{tabular}{l l r r r r c l}
\toprule
QueryProof vs & Metric & QueryProof & Baseline & Difference & Paired 95\% CI & Resolved & Verb \\
\midrule
32B direct & BTR (up) & 0.537 & 0.300 & $+0.237$ & $[+0.112, +0.375]$ & yes & outperforms \\
32B direct & FS share/task (down) & 0.163 & 0.613 & $-0.450$ & $[-0.575, -0.337]$ & yes & -- \\
32B direct & Cost/task (down) & 0.00081 & 0.00215 & $-0.00135$ & $[-0.00148, -0.00121]$ & yes & -- \\
32B direct & CPCA (down) & \$0.0017 & \$0.0058 & $-0.00414$ ($-71.0$\%) & $[-0.00752, -0.00250]$ & yes & -- \\
7B few-shot & BTR (up) & 0.537 & 0.275 & $+0.262$ & $[+0.137, +0.400]$ & yes & outperforms \\
7B few-shot & FS share/task (down) & 0.163 & 0.688 & $-0.525$ & $[-0.650, -0.400]$ & yes & -- \\
7B few-shot & Cost/task (down) & 0.00081 & 0.00060 & $+0.00020$ & $[+0.00007, +0.00037]$ & yes & -- \\
7B few-shot & CPCA (down) & \$0.0017 & \$0.0018 & $-0.00012$ ($-6.5$\%) & $[-0.00142, +0.00078]$ & no & -- \\
\bottomrule
\end{tabular}
\caption{Paired comparisons on the test split (80 tasks, 2000 resamples, seed 9901)}
\label{tab:t2_paired_test}
\end{table*}

%% file: tables/T3_by_category.tex
\begin{table}[tb]
\centering
\scriptsize
\begin{tabular}{lrrrrr}
\toprule
System & Std. & Amb. & Unans. & Drift & Adv. \\
\midrule
QueryProof routed & 0.500 & 0.250 & 0.438 & 1.000 & 1.000 \\
QueryProof base & 0.562 & 0.250 & 0.438 & 1.000 & 1.000 \\
32B direct & 0.500 & 0.000 & 0.000 & 0.000 & 1.000 \\
7B retrieval & 0.406 & 0.188 & 0.000 & 0.000 & 1.000 \\
7B few-shot & 0.438 & 0.000 & 0.000 & 0.000 & 1.000 \\
7B direct & 0.000 & 0.000 & 0.000 & 0.000 & 1.000 \\
\bottomrule
\end{tabular}
\caption{Business Truth Rate by task type, test split}
\label{tab:t3_by_category}
\end{table}

%% file: sections/08_limitations.tex
%% 08_limitations.tex
%%
%% Brief: docs/prd-publication.md section 7, row `08_limitations.tex`.
%% Task T7 in docs/planning/publication.md; depends on T1.
%%
%% Must contain docs/test_protocol.md section 9 in full, plus anything the run
%% surfaced. Every item was written down before the test run with the
%% instruction that it be stated whatever the numbers showed; deleting one is a
%% protocol violation, not an edit.
%%
%% Grouped rather than listed one paragraph per item, on external review: the
%% one-per-item form ran to twenty paragraphs and repeated the disclosure
%% framing in each. Nothing was dropped in the regrouping.
%%
%% Every number is a macro from macros/numbers.tex. Do not type a digit.

\section{Limitations}
\label{sec:limitations}

Every item here was pre-registered before the test run, with the instruction that
it be stated whatever the numbers showed. Three were added by the run and are
marked.

\paragraph{Two threats are serious enough to lead with.} The \emph{test split was
exposed} before the run: five hand-authored lexicon phrases matched the entire
test ambiguous stratum and could only have been written from the held-out split
(Section~\ref{sec:benchmark}). Removing them before any test-split model call
does not undo the exposure, because development decisions taken with that
knowledge cannot be reversed by deleting strings. Read the test figures as one
frozen evaluation with disclosed prior exposure, not as a clean confirmatory
test. Separately, the \emph{headline comparison is confounded}: QueryProof and
the 32B baseline differ in parameter count and in everything the scaffolding
provides, and the baseline is prompted directly and receives none of it. Nothing
here separates scale from scaffolding, part of the gap is information asymmetry
rather than reasoning, and no claim about model size should be drawn from these
numbers. The experiment that would separate them, the same scaffold over the 32B
model, needs a further evaluation of a spent split and is the first piece of
future work.

\paragraph{The evidence base is narrow.} \nTasksTotal{} tasks, two synthetic
domains generated from a single seed, one model family, one SQL dialect, and no
external benchmark. Sensitivity to the particular schema realisation is
unmeasured, transfer conclusions are specific to the one model family, and
nothing here demonstrates portability to Spider, BIRD or any other suite. The
warehouses are generated rather than sampled from production, so transfer to a
real warehouse with real schema debt is untested and is the obvious next
experiment. One person designed the warehouses, the task taxonomy, the semantic
layer, the deterministic rules, the labels and the evaluation harness;
co-design of that kind can favour the system in ways no single check detects.

\paragraph{The statistics are weaker than the intervals suggest.} The validation
split was evaluated six times during development, plus once more to select the
routing threshold, so every validation figure in this paper is an optimistic
estimate and appears only beside its test counterpart. The test split holds
\nTestTasks{} tasks drawn from \nFamiliesTest{} template families, so tasks are
not independent and the pre-registered task-level bootstrap treats them as though
they were. Resampling families instead widens both accuracy intervals to
include zero (Section~\ref{sec:results:cluster}), which is why the paper claims
the direction of the effect and not its magnitude. \emph{Added by the run:} the
cost-per-correct-answer comparison against the cost-matched baseline is
\resolvedCpcaSevenBFewShotTest{}, its interval spans zero, and on validation it
ran the other way. The confidence model is fitted on returned development
answers and evaluated on \nAnswersTest{} test answers, with generalisation
estimated leave-one-family-out because a row-wise estimate is roughly five times
more optimistic.

\paragraph{Several measurements are narrower than their names suggest.} A
verified answer is one that passed the deterministic checks of
Section~\ref{sec:method}; the term carries no correctness guarantee, and
\falseCleanDev{} of development answers and \falseCleanVal{} of validation
answers that passed them were wrong. An ambiguous task counts as correct when
the clarification contains a token distinguishing the competing interpretations,
which is checkable but does not establish that an analyst could act on it; no
human study was run. Result equivalence orders columns by content and excludes
headers, so a result with the right values under the wrong column roles can pass
where the column count still matches. Cost is an amortised local hardware rate
rather than a market price, so only the ratios between systems are meaningful.
Every system decodes at temperature zero, which removes sampling variance within
a configuration and says nothing about robustness across seeds or decoding
schemes.

\paragraph{Two strata do not support the claims they look like they support.}
The \nTestAdversarial{} adversarial tasks are passed by every system, so the
stratum confirms the safety property and ranks nothing; no comparative safety
claim is made. The \nTestAmbiguous{} ambiguous test tasks come from two template
families in one domain and exercise only the undefined-concept path, never
ambiguity between two defined metrics, which is the agent's primary mechanism,
so ambiguity results on test generalise poorly.

\paragraph{Labels rest on one reviewer.} The second pass was performed by the
original author, blind, and is a \adjudicationMeasure{} figure
(Table~\ref{tab:t7_adjudication}), not inter-annotator agreement. It bounds
instability in the labelling rule, does not remove single-reviewer bias, and its
disagreements all fell in one template family, so the effective number of
independent judgements is well below \adjudicated{}.

\paragraph{Two components are unresolved rather than understood.}
Post-training is untested by design: the gate was not met, with a pool of
\gatePostPool{} against a threshold of \gateThreshold{}
(Appendix~\ref{sec:gatea}), and whether execution-verified traces would help
remains open. \emph{Added by the run:} the routing threshold was selected on
validation and over-abstains on test
(Section~\ref{sec:results:routing}). It was not re-tuned, because that needs a
second evaluation of a spent split, so treat the operating point as an
illustration rather than a recommendation and select it on data from your own
distribution.

%% file: sections/09_conclusion.tex
%% 09_conclusion.tex
%%
%% Brief: docs/prd-publication.md section 7, row `09_conclusion.tex`. Target ~300 words.
%% Task T9 in docs/planning/publication.md; depends on T7, T8.
%%
%% Before drafting this file, read in order:
%%   1. docs/prd-publication.md sections 3-5  (claims, negative results, forbidden claims)
%%   2. this section's row in docs/prd-publication.md section 7
%%   3. reports/results_package/results.json
%% Nothing else is authoritative.
%%
%% Must contain:
%%   - What holds, what did not transfer, what a practitioner should take.
%%   - NO CLAIM ABSENT FROM SECTION 6.
%%
%% Every number is a macro from macros/numbers.tex. Do not type a digit.

\section{Conclusion}
\label{sec:conclusion}

This paper set out to test whether an analytics agent should be optimised for
reliable business answers rather than SQL syntax accuracy, and where that
reliability actually comes from.

What holds is the deterministic part. On a frozen split evaluated once, a
rule-gated 7B agent \verbBtrThirtyTwoBTest{} a direct-prompted 32B baseline on
task success by
\deltaBtrThirtyTwoBTest{} \deltaBtrThirtyTwoBTestCI{} at
\relCpcaThirtyTwoBTest{} lower cost per correct answer, and cuts false success
to \fsrRoutedTest{} of returned answers. The interval on that difference excludes
zero as pre-registered, but not once template families rather than tasks are
resampled (Section~\ref{sec:results:cluster}), so the ordering is better
established than the magnitude. It returns
\taxRoutedSilentWrong{} silent wrong answers on \nTestTasks{} tasks. The
separation comes almost entirely from the strata where the correct response is
not an answer, which is precisely where syntax-oriented evaluation looks away.

What did not transfer is the learned part. Routing helped on validation and
hurt on test by over-abstaining, and the fitted confidence model lost to the
heuristic it replaced on the same answers. Post-training was declined by a
pre-registered gate because the verification layer had already absorbed the
errors it would have targeted, leaving a pool of \gatePostPool{} against a
threshold of \gateThreshold{}; whether it would have helped remains open.

The practical implication is narrow, and applies to settings resembling the
warehouses and model family evaluated here. The measured gain is consistent with
the deterministic components being its main source: deciding behaviour by rule
against a semantic layer and a physical catalog, and checking results after
execution. Neither requires a weight update, a larger model, or a per-request
model judgement that cannot be audited, although both require the semantic layer
to be maintained. Attributing the gain to those components specifically would
need a component-wise ablation, which this paper does not report. The learned
component added on top is the part that failed to generalise, which suggests
building and measuring the deterministic layer first, and testing any confidence
model fitted on the answers it admits on a split never used to tune it.

%% file: appendix/a_scoring_corrections.tex
%% a_scoring_corrections.tex
%%
%% Brief: docs/prd-publication.md section 7, row `a_scoring_corrections.tex`. ~400 words.
%% Task T10 in docs/planning/publication.md; depends on T1.
%%
%% Must contain:
%%   - T6. Every correction, and whether it moved a behavioural metric.
%%   - The Phase 6 pair forced WRB v0.1.0 -> v0.2.0; the later ones did not.
%%
%% Every number is a macro from macros/numbers.tex. Do not type a digit.

\section{Scoring corrections}
\label{sec:corrections}

Six defects were found in the scoring harness over the life of the project,
four of them after results had already been reported internally. All are listed
below, with whether each moved a behavioural metric, because a reader has no way
to audit a scorer they cannot see and every undisclosed correction is a reason
to distrust the disclosed ones.
Table~\ref{tab:t6_scoring_corrections} is the full list.

Two patterns are worth naming.

\paragraph{Denominators were the recurring defect.} Three of the six
corrections are denominator errors: False Success Rate was computed over all
tasks rather than over outputs labelled \textsc{answer}; cost per correct answer
was computed as cost per \emph{success}, which counts clarifications,
abstentions and refusals and therefore flatters a system that earns its score by
declining to answer; and expected calibration error pooled every task rather
than every returned answer. Each was a plausible reading of an underspecified
metric name, and each changed the reported figure materially. This is why the
metric definitions in this paper state their denominator explicitly, and why
Section~\ref{sec:results} restates the pool whenever a rate appears.

\paragraph{Only the first two changed a number.} The two corrections found
during agent development changed reported values and forced the benchmark
version bump to v\wrbVersion{}, so every figure in this paper is stated against
that version. Each later correction was verified to leave task success, false
success, coverage and answer accuracy bit-identical before and after, which is
why none required a further version bump. That verification is not an assurance
offered on trust: it is a re-scoring of the stored runs, reproducible from the
repository.

The one correction that could have invalidated the pre-registration, the
clarification-coverage rule changed during the measurement close-out, was made
\emph{before} the test run and applied to a secondary metric. The
superseded rule is retained under a separate name so that the earlier figures
remain interpretable rather than being silently overwritten.
\input{tables/T6_scoring_corrections}

%% file: tables/T6_scoring_corrections.tex
\begin{table*}[tb]
\centering
\footnotesize
\begin{tabularx}{\textwidth}{r X X X c}
\toprule
Phase & Change & Before & After & Moved metric? \\
\midrule
6 & FSR denominator & incorrect answers / all tasks & incorrect answers / outputs labelled ANSWER (project-definition \S{}5) & yes \\
6 & result equivalence hashed column names & column aliases part of the contract & columns ordered by content, headers excluded; column count still checked & yes \\
10 & clarification coverage rule & any shared content token counted an interpretation as covered & a token unique to that interpretation is required; superseded rule kept as clarification\_topicality & no \\
10 & cost measurement sealed into the run record & execution time re-measured at every rescore, drifting 6e-6 USD & measured once, quantized to 1e-6 USD, frozen as data & no \\
10 & CPCA separated from cost-per-success & total cost / all correct behaviors, labelled CPCA & cost on answerable tasks / correct returned answers (\S{}5); cost-per-success reported separately & no \\
11 & ECE scoped to returned answers & ECE over every task, mixing in abstentions and clarifications & \S{}5 scoping to returned answers; all-task figure retained separately & no \\
\bottomrule
\end{tabularx}
\caption{Every scoring correction, and whether it moved a behavioural metric}
\label{tab:t6_scoring_corrections}
\end{table*}

%% file: appendix/b_reproduction.tex
%% b_reproduction.tex
%%
%% Brief: docs/prd-publication.md section 7, row `b_reproduction.tex`. ~300 words.
%% Task T10 in docs/planning/publication.md; depends on T1.
%%
%% Must contain:
%%   - ./scripts/reproduce.sh; one command per table and figure.
%%   - The sealed-cost rationale; artifact hashes.
%%
%% Every number is a macro from macros/numbers.tex. Do not type a digit.

\section{Reproduction}
\label{sec:reproduction}

Every number in this paper can be regenerated from the repository,
\url{https://github.com/k-w-lee/query_proof}, with one command and no model
calls:

\begin{quote}
\texttt{./scripts/reproduce.sh}
\end{quote}

The raw outputs of all six systems are committed, so scoring, aggregation, the
paired bootstrap, the tables and the figures are all replayable. The script
rebuilds the warehouse from its pinned seed, runs the test suite (including the
guard that every golden SQL program passes the agent's own validator, and the
guard against test-derived phrases in the lexicon), scores every run,
recomputes the paired comparisons, regenerates the Gate A evidence, and
consolidates everything into the results package. Regenerating the model
outputs themselves is a separate, longer path and requires the same hardware;
the frozen test split must never be re-run.

\paragraph{One source, hashed.} Tables and figures read a single consolidated
artifact, \texttt{results.json}, which records a content hash for each of the
\nHashedInputs{} inputs it consumed, together with the commit that produced the
runs (\runsCommit{}) and the commit of the analysis code (\analysisCommit{}).
A figure therefore cannot disagree with the table beside it, and a stale
artifact is detectable rather than merely suspected. Every figure carries the
runs commit in its footer.

\paragraph{Costs are sealed, not re-measured.} The execution component of cost
is a wall-clock measurement, so recomputing it at each rescore made reported
cost drift with the evaluator's own thread contention, above the reporting
quantum on an \nTestTasks{}-task split. Cost is therefore measured once,
serially, quantised, and frozen into the run record as data; scoring reads the
sealed value. Reports are byte-identical across repeated rescores as a result,
which is what makes the claim of reproducibility checkable rather than
aspirational.

\paragraph{Replay is not regeneration.} Two different claims are worth
separating. \emph{Replay} reproducibility is what the command above gives:
scoring, aggregation, the paired tests, the tables and the figures all rebuild
byte-identically from the committed model outputs, on any machine, with no model
calls. \emph{Generation} reproducibility, regenerating those model outputs, is
weaker: it needs the same quantised weights, the same inference backend and
comparable hardware, and greedy decoding on a different backend may still differ.
Every number in this paper is replay-reproducible. None of them should be assumed
bit-reproducible from a fresh generation on different hardware.

\paragraph{The manuscript is a build product too.} The tables and figures in
this paper are generated by \texttt{make -C paper all} and are never
hand-edited, and every headline number in the prose is a macro generated from
\texttt{results.json} rather than typed. A figure that disagreed with its own
source would fail to build, not merely fail review.

%% file: appendix/c_supporting_floats.tex
%% c_supporting_floats.tex
%%
%% Supporting floats moved out of the body for length. NOT a demotion of the
%% evidence: every claim these support is stated with its numbers in the body,
%% and each is cross-referenced from the sentence that relies on it. The
%% pre-registered set (test_protocol.md section 8) is unchanged -- this is
%% placement only.
%%
%% Every number is a macro from macros/numbers.tex. Do not type a digit.

\section{Supporting tables and figures}
\label{sec:supporting}

\begin{figure*}[tb]
\centering
\includegraphics[width=\textwidth]{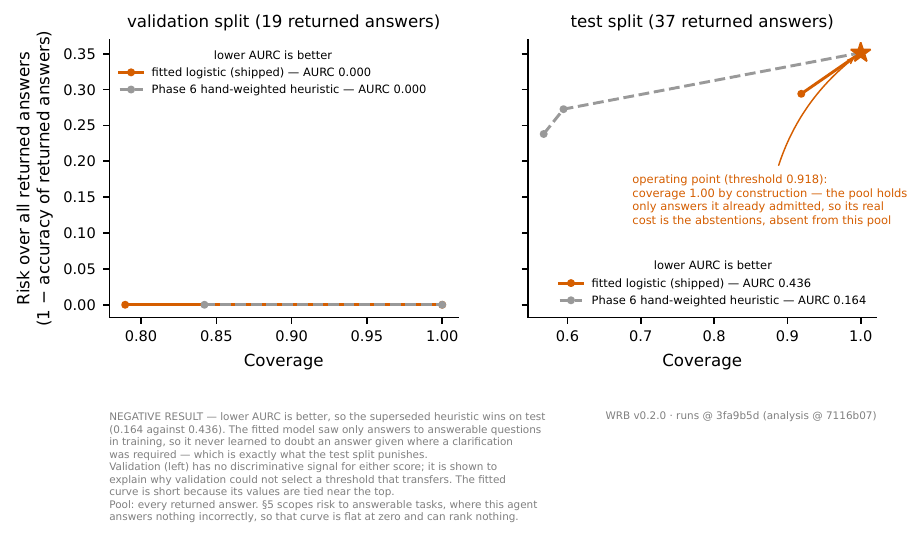}
\caption{F2. \textbf{Negative result.} Risk--coverage for the fitted confidence
model against the Phase~6 heuristic, on the same answer pool. Lower AURC is
better.}
\label{fig:f2}
\end{figure*}

\begin{figure*}[tb]
\centering
\includegraphics{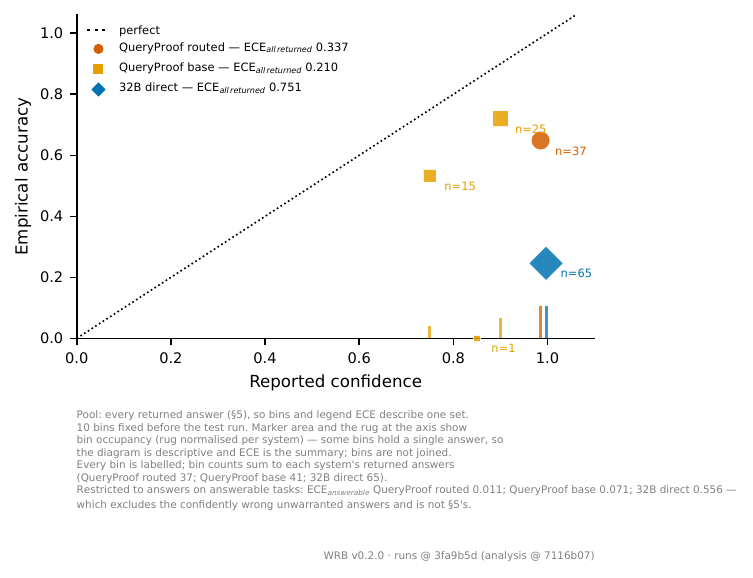}
\caption{F3. Reliability diagram over \nAnswersTest{} returned answers, with the
ten bins fixed before the test run. Marker size follows bin occupancy.}
\label{fig:f3}
\end{figure*}

This appendix holds the wide floats the body cites but does not need inline. None of
it is optional reading: Table~\ref{tab:t5a_routing_ablation} is the evidence for
the routing negative result, and Figures~\ref{fig:f2} and~\ref{fig:f3} are the
evidence that the fitted confidence model loses to the heuristic it replaced.

\begin{figure*}[tb]
\centering
\includegraphics[width=\textwidth]{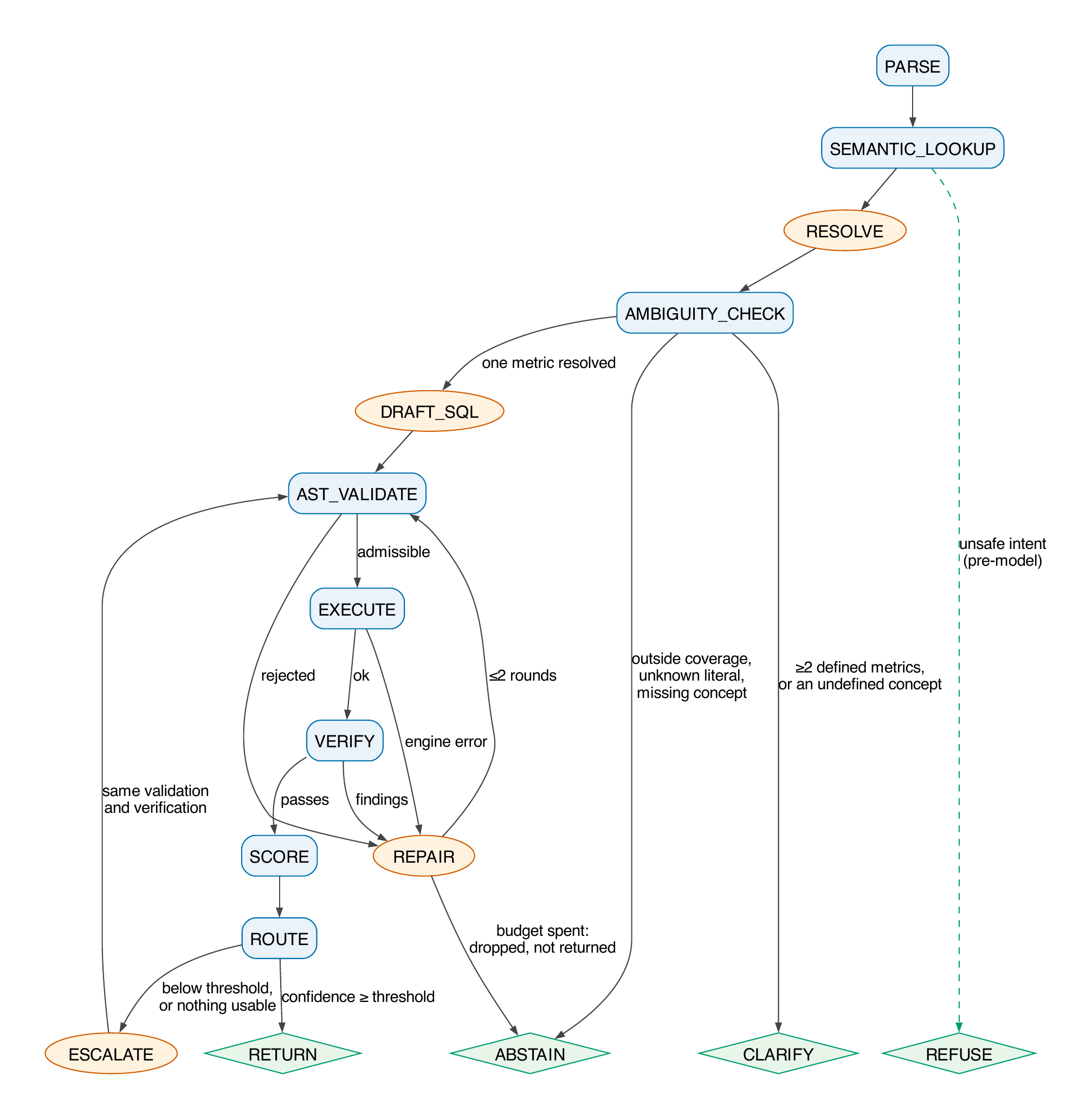}
\caption{The complete agent state machine, of which
Figure~\ref{fig:policy} is the decision ordering. \emph{Methods figure}, not a
result. Shape carries authority: box = rule, ellipse = model call, diamond =
terminal behaviour. What each state does is given in
Section~\ref{sec:method}; labels are state names only, so the diagram stays
legible at page scale.}
\label{fig:f0}
\end{figure*}

\begin{table}[tb]
\centering
\footnotesize
\begin{tabular}{@{}ll@{}}
\toprule
Concern & Decided by \\
\midrule
Is the request safe? & rule \\
What period is meant? & rule \\
Does the warehouse cover it? & rule \\
Is that a real dimension value? & rule \\
Has the term one definition? & rule \\
Which candidate metric is meant? & \textbf{model}, re-checked \\
What is missing from the schema? & \textbf{model}, re-checked \\
What SQL computes it? & \textbf{model} \\
Is the SQL admissible? & rule \\
Is the result trustworthy? & rule \\
How likely is the answer right? & model fitted on dev \\
Return, escalate or abstain? & validation threshold \\
\bottomrule
\end{tabular}
\caption{Where authority sits. \emph{Methods table}, not a result: like
Figure~\ref{fig:policy} it describes the system and carries no claim. Every model output is re-validated
against the semantic layer or the catalog before it can change the outcome.}
\label{tab:authority}
\end{table}

\input{tables/T8_outcome_reconciliation}
\input{tables/T9_cluster_robustness}
\input{tables/T4_validation_vs_test}
\input{tables/T5a_routing_ablation}
\input{tables/T5b_defect_sweep}

%% file: tables/T8_outcome_reconciliation.tex
\begin{table*}[tb]
\centering
\scriptsize
\begin{tabular}{lrrrrrr}
\toprule
Outcome & QueryProof routed & QueryProof base & 32B direct & 7B retrieval & 7B few-shot & 7B direct \\
\midrule
\textbf{Answer returned} &  &  &  &  &  &  \\
\quad correct answer & 24 & 26 & 16 & 13 & 14 & 0 \\
\quad wrong number, answerable task & 0 & 2 & 4 & 4 & 13 & 18 \\
\quad answered an ambiguous question & 4 & 4 & 16 & 13 & 16 & 16 \\
\quad answered an unanswerable question & 9 & 9 & 13 & 14 & 13 & 16 \\
\quad SQL failed to execute & 0 & 0 & 16 & 23 & 13 & 22 \\
\textbf{No answer returned} &  &  &  &  &  &  \\
\quad correct clarification, abstention or refusal & 19 & 19 & 8 & 11 & 8 & 8 \\
\quad no answer, clarification required & 8 & 8 & 0 & 0 & 0 & 0 \\
\quad no answer, abstention required & 0 & 0 & 3 & 2 & 3 & 0 \\
\quad declined an answerable question & 12 & 8 & 1 & 0 & 0 & 0 \\
\quad clarified an unambiguous question & 4 & 4 & 3 & 0 & 0 & 0 \\
\textbf{Total} & 80 & 80 & 80 & 80 & 80 & 80 \\
FSR numerator (rows 2-5) & 13 & 15 & 49 & 54 & 55 & 72 \\
FSR denominator (rows 1-5) & 37 & 41 & 65 & 67 & 69 & 72 \\
\bottomrule
\end{tabular}
\caption{Every test task in one mutually exclusive outcome (80 tasks)}
\label{tab:t8_outcome_reconciliation}
\end{table*}

%% file: tables/T9_cluster_robustness.tex
\begin{table*}[tb]
\centering
\footnotesize
\begin{tabular}{l l r r r c}
\toprule
QueryProof vs & Resampling unit & n & Difference & 95\% CI & Excludes zero \\
\midrule
32B direct & task (pre-registered) & 80 tasks & $+0.237$ & $[+0.112, +0.375]$ & yes \\
32B direct & template family (exploratory) & 10 families & $+0.237$ & $[-0.125, +0.562]$ & \textbf{no} \\
7B few-shot & task (pre-registered) & 80 tasks & $+0.262$ & $[+0.137, +0.400]$ & yes \\
7B few-shot & template family (exploratory) & 10 families & $+0.263$ & $[-0.075, +0.575]$ & \textbf{no} \\
\bottomrule
\end{tabular}
\caption{Business Truth Rate difference, resampling tasks against families}
\label{tab:t9_cluster_robustness}
\end{table*}

%% file: tables/T4_validation_vs_test.tex
\begin{table*}[tb]
\centering
\small
\begin{tabular}{lrrrrrrr}
\toprule
System & BTR val & BTR test & $\Delta$ BTR & FSR val & FSR test & ECE val & ECE test \\
\midrule
QueryProof routed & 0.738 & 0.537 & $-0.200$ & 0.000 & 0.351 & 0.023 & 0.337 \\
QueryProof base & 0.725 & 0.562 & $-0.162$ & 0.308 & 0.366 & 0.227 & 0.210 \\
32B direct & 0.475 & 0.300 & $-0.175$ & 0.537 & 0.754 & 0.537 & 0.751 \\
7B retrieval & 0.250 & 0.300 & $+0.050$ & 0.793 & 0.806 & 0.793 & 0.806 \\
7B few-shot & 0.312 & 0.275 & $-0.037$ & 0.741 & 0.797 & 0.641 & 0.697 \\
7B direct & 0.163 & 0.100 & $-0.062$ & 0.919 & 1.000 & 0.919 & 1.000 \\
\bottomrule
\end{tabular}
\caption{Validation against the frozen test split}
\label{tab:t4_validation_vs_test}
\end{table*}

%% file: tables/T5a_routing_ablation.tex
\begin{table*}[tb]
\centering
\footnotesize
\begin{tabular}{lrrrrr}
\toprule
System & BTR val & BTR test & Cov. & Ans. acc. & FSR \\
\midrule
QueryProof base & 0.725 & 0.562 & 0.700 & 0.929 & 0.366 \\
QueryProof routed & 0.738 & 0.537 & 0.600 & 1.000 & 0.351 \\
\bottomrule
\end{tabular}
\caption{Routing ablation: calibrated abstention and escalation on/off}
\label{tab:t5a_routing_ablation}
\end{table*}

%% file: tables/T5b_defect_sweep.tex
\begin{table*}[tb]
\centering
\footnotesize
\begin{tabular}{lrrrrr}
\toprule
Stage & Tasks & Fail. & Agent & Model pool & Threshold \\
\midrule
before the Phase 9 sweep & 240 & 33 & 21 & 12 & 100 \\
after the Phase 9 sweep & 240 & 28 & 8 & 20 & 100 \\
\bottomrule
\end{tabular}
\caption{Phase 9 deterministic defect sweep, dev split}
\label{tab:t5b_defect_sweep}
\end{table*}

%% file: appendix/d_gate_a.tex
%% 07_gate_a.tex
%%
%% Brief: docs/prd-publication.md section 7, row `07_gate_a.tex`. Target ~600 words.
%% Task T4 in docs/planning/publication.md; depends on T1.
%%
%% Before drafting this file, read in order:
%%   1. docs/prd-publication.md sections 3-5  (claims, negative results, forbidden claims)
%%   2. this section's row in docs/prd-publication.md section 7
%%   3. reports/results_package/results.json
%% Nothing else is authoritative.
%%
%% Must contain:
%%   - The pre-registered rule; the root-cause analysis.
%%   - Pool 12 -> 20 against a threshold of 100.
%%   - The reversal conditions, declared in advance.
%%   - Frame as a finding: the verification layer captured the headroom that
%%     post-training would have targeted.
%%
%% H4 is reported as DECLINED WITH EVIDENCE, never as a negative result and
%% never silently dropped.
%%
%% Every number is a macro from macros/numbers.tex. Do not type a digit.

\section{Gate A: the post-training decision}
\label{sec:gatea}

This project planned to post-train the 7B model on execution-verified traces.
It did not, and the reason is a finding rather than an omission. A pre-registered gate
was not met, so the component was not built. The gate evidence is reported below
so the decision can be checked rather than taken on trust.

\subsection{The rule, fixed in advance}

The project definition, written before any agent existed, permitted
post-training only if three conditions held together: the benchmark and scoring
harness were stable, all required baselines had completed, and at least
\gateThreshold{} verified training examples represented recurring errors that
prompting or retrieval did not solve. Otherwise the remaining budget was to go
to benchmark quality, verification, calibration and failure analysis. The
threshold was fixed before any of the data existed.

The first two conditions were met. The third was not.

\subsection{Sizing the pool}

Training data may only be drawn from the development split, so the pool is
bounded by the agent's residual failures on the \nDevSplit{} development tasks.
Counting those failures is the wrong measurement, because a failure caused by a
bug in the deterministic layer is fixed by editing that layer; putting it in a
training set would teach the model to compensate for a bug, producing a worse
system that is also harder to audit.

Every failure was therefore assigned a root cause from trace evidence, and only
the \emph{model-capability} class can be addressed by training.
Table~\ref{tab:t5b_defect_sweep} reports the sweep. Before the deterministic
fixes, \gatePreFailures{} dev failures decomposed into \gatePreDefects{} agent
defects and a model-capability pool of \gatePrePool{}, against a threshold of
\gateThreshold{}. The condition failed by a factor of eight.

The strongest single piece of evidence was in the agent, not the model: running
each answerable task's \emph{own golden SQL} through the agent's static
validator rejected queries from two template families, because an invalid-join
rule read "no direct key between these tables" as "these tables may never
co-occur". That is not evidence about the model at all, but evidence of a missing
regression test, which now exists.

\subsection{The reversal condition was tested, and held}

The decision was made falsifiable in advance. Three conditions were declared
that would reverse it: that fixing the deterministic defects would expose more
than \gateThreshold{} model failures on dev; that expanding the benchmark toward
its stretch size would produce that many coherent model-capability failures; or
that calibration would fail to recover coverage with the trace evidence blaming
SQL generation quality rather than conservative thresholds.

The first was then actually tested. All four defect classes were fixed and the
agent re-run on dev. Failures fell from \gatePreFailures{} to
\gatePostFailures{} and agent defects from \gatePreDefects{} to
\gatePostDefects{}, while the model-capability pool \emph{grew} from
\gatePrePool{} to \gatePostPool{}, exactly the direction predicted, since
tasks that previously died on a bug now reach the model and can fail for genuine
reasons. At \gatePostPool{} the pool remains five times below the threshold, and
the decision stands.

\subsection{What this means}

The supported reading is not that post-training would not have helped. It is
narrower: \emph{on this benchmark, a deterministic
verification layer captured most of the available reliability headroom, leaving
too small a coherent error set to justify post-training.} The agent resolved the
large majority of every baseline's development failures with no weight update,
and those failures were overwhelmingly cases where the system should have
clarified or abstained, which a rule reading the semantic layer decides directly
from the metric definitions. How a fine-tuned model would have handled them was
not measured, and is not asserted here.
The same pattern is visible on the frozen test split in
Table~\ref{tab:t3_by_category}, where the baselines score zero on the
unanswerable and schema-drift strata.

The hypothesis that execution-verified traces improve a small analytics model is
therefore \textbf{untested by design} and reported as such. It is not a negative
result: the experiment was not run, and no outcome is implied for it. What can
be stated is why it was not run, with the number that decided it and the
conditions that would have reversed the decision.

%% file: appendix/e_metrics.tex
%% e_metrics.tex
%%
%% Added on external review: the paper reported six metrics whose denominators
%% differ and stated them only in prose and table captions. Placed in the
%% appendix rather than the body because the same review asks that detailed
%% metric definitions live here, with the body carrying the scope in words.
%%
%% These are definitions, not results: no number appears.

\section{Metric definitions}
\label{sec:metrics}

Let $T$ be the tasks in scope and $n = |T|$. For a task $t$, the benchmark fixes
a required behaviour $b^{*}(t) \in \{\textsc{answer}, \textsc{clarify},
\textsc{abstain}, \textsc{refuse}\}$, and a system emits a behaviour $b(t)$ and,
when $b(t) = \textsc{answer}$, a result. Write $A = \{t : b(t) =
\textsc{answer}\}$ for the tasks the system answered and $Q = \{t : b^{*}(t) =
\textsc{answer}\}$ for the answerable tasks. Let $\mathrm{ok}(t)$ be $1$ when the
task is scored correct: the required behaviour was produced, and for $t \in Q$
the returned result is equivalent to an accepted result under the contract of
Section~\ref{sec:benchmark}.

The denominators differ between metrics, which is the point of stating them:

\begin{align}
\text{BTR} &= \tfrac{1}{n} \textstyle\sum_{t \in T} \mathrm{ok}(t) \\[2pt]
\text{FSR} &= \tfrac{1}{|A|} \textstyle\sum_{t \in A} \bigl(1 - \mathrm{ok}(t)\bigr) \\[2pt]
\text{coverage} &= |A \cap Q| \,/\, |Q| \\[2pt]
\text{answer accuracy} &= \tfrac{1}{|A \cap Q|} \textstyle\sum_{t \in A \cap Q} \mathrm{ok}(t) \\[2pt]
\text{CPCA} &= \textstyle\sum_{t \in Q} c(t) \;\Big/\; \textstyle\sum_{t \in A \cap Q} \mathrm{ok}(t)
\end{align}

where $c(t)$ is the variable cost of task $t$ under Section~\ref{sec:setup}.
Business Truth Rate is over \emph{all} tasks in scope and rewards a correct
clarification, abstention or refusal exactly as it rewards a correct answer.
False Success Rate is over \emph{answers only}: it is silent about tasks where
the system returned nothing, which is why it must be read beside coverage.
Coverage and answer accuracy are over \emph{answerable} tasks, so a system that
answers everything scores coverage $1$ regardless of what it answers. CPCA is a
ratio of sums, not a mean of per-task ratios, and is therefore undefined for a
system with no correct answer.

\paragraph{Two quantities that are not FSR.} A paired test needs one value per
task over a fixed task list, so the comparison in
Table~\ref{tab:t2_paired_test} uses the \emph{false-success share per task},
\begin{equation}
\text{FS share} = \tfrac{1}{n} \textstyle\sum_{t \in T} \mathbf{1}\!\left[t \in A \wedge \mathrm{ok}(t) = 0\right],
\end{equation}
whose denominator is $n$ rather than $|A|$. It is smaller than FSR whenever the
system declines some tasks. Cost per \emph{success} divides total cost by every
correct behaviour, including correct abstentions, and so flatters a system that
earns its score by declining; it is reported separately from CPCA and is never
the headline.

\paragraph{Calibration.} Expected calibration error uses ten equal-width bins
fixed before the test run, over the pool $A$ of returned answers:
\begin{equation}
\text{ECE} = \textstyle\sum_{i=1}^{10} \frac{|B_i|}{|A|}
             \bigl|\, \mathrm{acc}(B_i) - \mathrm{conf}(B_i) \,\bigr|,
\end{equation}
with $B_i$ the answers whose confidence falls in bin $i$. The narrower pool $A
\cap Q$ is degenerate on the test split, as Section~\ref{sec:results} explains.

\paragraph{Paired comparison.} For systems $x$ and $y$ evaluated on the same
tasks and a per-task metric $m$, the reported quantity is the mean paired
difference $\bar{d} = \tfrac{1}{n}\sum_{t}\left(m_x(t) - m_y(t)\right)$, with a
percentile interval from \pairedResamplesTest{} bootstrap resamples of the task list under a
fixed seed. The pre-registered rule permits \emph{outperforms} only when that
interval excludes zero. Section~\ref{sec:results:cluster} reports the same
difference resampling template families instead of tasks.

%% file: main.bbl
\begin{thebibliography}{26}
\providecommand{\natexlab}[1]{#1}

\bibitem[{Bhaskar et~al.(2023)Bhaskar, Tomar, Sathe, and
  Sarawagi}]{bhaskar2023benchmarking}
Adithya Bhaskar, Tushar Tomar, Ashutosh Sathe, and Sunita Sarawagi. 2023.
\newblock \href {https://doi.org/10.18653/v1/2023.emnlp-main.436} {Benchmarking
  and improving text-to-{SQL} generation under ambiguity}.
\newblock In \emph{Proceedings of the 2023 Conference on Empirical Methods in
  Natural Language Processing}, pages 7053--7074, Singapore. Association for
  Computational Linguistics.

\bibitem[{{Cube Dev, Inc.}(2026)}]{cube2026semanticlayer}
{Cube Dev, Inc.} 2026.
\newblock \href {https://docs.cube.dev/introduction} {{Cube}: Introduction}.
\newblock Cube documentation.
\newblock Accessed 2026-08-09.

\bibitem[{{dbt Labs}(2026{\natexlab{a}})}]{dbt2026metricflow}
{dbt Labs}. 2026{\natexlab{a}}.
\newblock \href {https://docs.getdbt.com/docs/build/about-metricflow} {About
  {MetricFlow}}.
\newblock dbt Developer Hub documentation.
\newblock Accessed 2026-08-09.

\bibitem[{{dbt Labs}(2026{\natexlab{b}})}]{dbt2026semanticlayer}
{dbt Labs}. 2026{\natexlab{b}}.
\newblock \href {https://docs.getdbt.com/docs/use-dbt-semantic-layer/dbt-sl}
  {dbt semantic layer}.
\newblock dbt Developer Hub documentation.
\newblock Accessed 2026-08-09.

\bibitem[{Efron and Tibshirani(1993)}]{efron1993bootstrap}
Bradley Efron and Robert~J. Tibshirani. 1993.
\newblock \href {https://doi.org/10.1201/9780429246593} {\emph{An Introduction
  to the Bootstrap}}.
\newblock Number~57 in Monographs on Statistics and Applied Probability.
  Chapman \& Hall/CRC, New York.

\bibitem[{El-Yaniv and Wiener(2010)}]{elyaniv2010foundations}
Ran El-Yaniv and Yair Wiener. 2010.
\newblock \href {https://www.jmlr.org/papers/v11/el-yaniv10a.html} {On the
  foundations of noise-free selective classification}.
\newblock \emph{Journal of Machine Learning Research}, 11(53):1605--1641.

\bibitem[{Gao et~al.(2024)Gao, Wang, Li, Sun, Qian, Ding, and
  Zhou}]{gao2024dailsql}
Dawei Gao, Haibin Wang, Yaliang Li, Xiuyu Sun, Yichen Qian, Bolin Ding, and
  Jingren Zhou. 2024.
\newblock \href {https://doi.org/10.14778/3641204.3641221} {Text-to-{SQL}
  empowered by large language models: A benchmark evaluation}.
\newblock \emph{Proceedings of the VLDB Endowment}, 17(5):1132--1145.

\bibitem[{Geifman and El-Yaniv(2017)}]{geifman2017selective}
Yonatan Geifman and Ran El-Yaniv. 2017.
\newblock \href {https://arxiv.org/abs/1705.08500} {Selective classification
  for deep neural networks}.
\newblock In \emph{Advances in Neural Information Processing Systems 30 (NIPS
  2017)}.

\bibitem[{{Google Cloud}(2026)}]{looker2026lookml}
{Google Cloud}. 2026.
\newblock \href {https://cloud.google.com/looker/docs/what-is-lookml}
  {Introduction to {LookML}}.
\newblock Looker documentation.
\newblock Accessed 2026-08-09.

\bibitem[{Guo et~al.(2017)Guo, Pleiss, Sun, and
  Weinberger}]{guo2017calibration}
Chuan Guo, Geoff Pleiss, Yu~Sun, and Kilian~Q. Weinberger. 2017.
\newblock \href {https://arxiv.org/abs/1706.04599} {On calibration of modern
  neural networks}.
\newblock In \emph{Proceedings of the 34th International Conference on Machine
  Learning}, volume~70 of \emph{Proceedings of Machine Learning Research},
  pages 1321--1330. PMLR.

\bibitem[{Hui et~al.(2024)Hui, Yang, Cui, Yang, Liu, Zhang, Liu, Zhang, Yu,
  Dang, Yang, Men, Huang, Ren, Ren, Zhou, and Lin}]{hui2024qwen25coder}
Binyuan Hui, Jian Yang, Zeyu Cui, Jiaxi Yang, Dayiheng Liu, Lei Zhang, Tianyu
  Liu, Jiajun Zhang, Bowen Yu, Kai Dang, An~Yang, Rui Men, Fei Huang, Xingzhang
  Ren, Xuancheng Ren, Jingren Zhou, and Junyang Lin. 2024.
\newblock \href {https://arxiv.org/abs/2409.12186} {{Qwen2.5-Coder} technical
  report}.
\newblock \emph{arXiv preprint arXiv:2409.12186}.

\bibitem[{Kadavath et~al.(2022)Kadavath, Conerly, Askell, Henighan, Drain,
  Perez, Schiefer, Hatfield-Dodds, DasSarma, Tran-Johnson, Johnston, El-Showk,
  Jones, Elhage, Hume, Chen, Bai, Bowman, Fort, Ganguli, Hernandez, Jacobson,
  Kernion, Kravec, Lovitt, Ndousse, Olsson, Ringer, Amodei, Brown, Clark,
  Joseph, Mann, McCandlish, Olah, and Kaplan}]{kadavath2022language}
Saurav Kadavath, Tom Conerly, Amanda Askell, Tom Henighan, Dawn Drain, Ethan
  Perez, Nicholas Schiefer, Zac Hatfield-Dodds, Nova DasSarma, Eli
  Tran-Johnson, Scott Johnston, Sheer El-Showk, Andy Jones, Nelson Elhage,
  Tristan Hume, Anna Chen, Yuntao Bai, Sam Bowman, Stanislav Fort, and 17
  others. 2022.
\newblock \href {https://arxiv.org/abs/2207.05221} {Language models (mostly)
  know what they know}.
\newblock \emph{arXiv preprint arXiv:2207.05221}.

\bibitem[{Lee et~al.(2024)Lee, Chay, Cho, and Choi}]{lee2024trustsql}
Gyubok Lee, Woosog Chay, Seonhee Cho, and Edward Choi. 2024.
\newblock \href {https://arxiv.org/abs/2403.15879} {{TrustSQL}: Benchmarking
  text-to-{SQL} reliability with penalty-based scoring}.
\newblock \emph{arXiv preprint arXiv:2403.15879}.

\bibitem[{Lei et~al.(2025)Lei, Chen, Ye, Cao, Shin, Su, Suo, Gao, Hu, Yin,
  Zhong, Xiong, Sun, Liu, Wang, and Yu}]{lei2025spider2}
Fangyu Lei, Jixuan Chen, Yuxiao Ye, Ruisheng Cao, Dongchan Shin, Hongjin Su,
  Zhaoqing Suo, Hongcheng Gao, Wenjing Hu, Pengcheng Yin, Victor Zhong, Caiming
  Xiong, Ruoxi Sun, Qian Liu, Sida~I. Wang, and Tao Yu. 2025.
\newblock \href {https://arxiv.org/abs/2411.07763} {Spider 2.0: Evaluating
  language models on real-world enterprise text-to-{SQL} workflows}.
\newblock In \emph{The Thirteenth International Conference on Learning
  Representations (ICLR)}.

\bibitem[{Li et~al.(2023)Li, Hui, Qu, Yang, Li, Li, Wang, Qin, Geng, Huo, Zhou,
  Ma, Li, Chang, Huang, Cheng, and Li}]{li2023bird}
Jinyang Li, Binyuan Hui, Ge~Qu, Jiaxi Yang, Binhua Li, Bowen Li, Bailin Wang,
  Bowen Qin, Ruiying Geng, Nan Huo, Xuanhe Zhou, Chenhao Ma, Guoliang Li, Kevin
  C.~C. Chang, Fei Huang, Reynold Cheng, and Yongbin Li. 2023.
\newblock \href {https://arxiv.org/abs/2305.03111} {Can {LLM} already serve as
  a database interface? a {BI}g bench for large-scale database grounded
  text-to-{SQL}s}.
\newblock In \emph{Advances in Neural Information Processing Systems 36
  (NeurIPS 2023) Track on Datasets and Benchmarks}.

\bibitem[{Lin et~al.(2022)Lin, Hilton, and Evans}]{lin2022teaching}
Stephanie Lin, Jacob Hilton, and Owain Evans. 2022.
\newblock \href {https://arxiv.org/abs/2205.14334} {Teaching models to express
  their uncertainty in words}.
\newblock \emph{Transactions on Machine Learning Research}.

\bibitem[{{Malloy Data}(2026)}]{malloy2026}
{Malloy Data}. 2026.
\newblock \href {https://www.malloydata.dev/} {{Malloy}: A modern open source
  language for analyzing, transforming, and modeling data}.
\newblock Project website and documentation.
\newblock Accessed 2026-08-09.

\bibitem[{Naeini et~al.(2015)Naeini, Cooper, and
  Hauskrecht}]{naeini2015obtaining}
Mahdi~Pakdaman Naeini, Gregory~F. Cooper, and Milos Hauskrecht. 2015.
\newblock \href {https://doi.org/10.1609/aaai.v29i1.9602} {Obtaining well
  calibrated probabilities using bayesian binning}.
\newblock In \emph{Proceedings of the Twenty-Ninth AAAI Conference on
  Artificial Intelligence (AAAI)}, volume~29, pages 2901--2907. AAAI Press.

\bibitem[{Nosek et~al.(2018)Nosek, Ebersole, DeHaven, and
  Mellor}]{nosek2018preregistration}
Brian~A. Nosek, Charles~R. Ebersole, Alexander~C. DeHaven, and David~T. Mellor.
  2018.
\newblock \href {https://doi.org/10.1073/pnas.1708274114} {The preregistration
  revolution}.
\newblock \emph{Proceedings of the National Academy of Sciences},
  115(11):2600--2606.

\bibitem[{Pourreza and Rafiei(2023)}]{pourreza2023dinsql}
Mohammadreza Pourreza and Davood Rafiei. 2023.
\newblock \href {https://arxiv.org/abs/2304.11015} {{DIN}-{SQL}: Decomposed
  in-context learning of text-to-{SQL} with self-correction}.
\newblock In \emph{Advances in Neural Information Processing Systems 36
  (NeurIPS 2023)}.

\bibitem[{Raasveldt and M{\"u}hleisen(2019)}]{raasveldt2019duckdb}
Mark Raasveldt and Hannes M{\"u}hleisen. 2019.
\newblock \href {https://doi.org/10.1145/3299869.3320212} {{DuckDB}: an
  embeddable analytical database}.
\newblock In \emph{Proceedings of the 2019 International Conference on
  Management of Data (SIGMOD '19)}, pages 1981--1984, Amsterdam, Netherlands.
  ACM.

\bibitem[{Wang et~al.(2023)Wang, Gao, Li, and Lou}]{wang2023know}
Bing Wang, Yan Gao, Zhoujun Li, and Jian-Guang Lou. 2023.
\newblock \href {https://doi.org/10.18653/v1/2023.findings-acl.352} {Know what
  {I} don{'}t know: Handling ambiguous and unknown questions for
  text-to-{SQL}}.
\newblock In \emph{Findings of the Association for Computational Linguistics:
  ACL 2023}, pages 5701--5714, Toronto, Canada. Association for Computational
  Linguistics.

\bibitem[{Wang et~al.(2018)Wang, Tatwawadi, Brockschmidt, Huang, Mao, Polozov,
  and Singh}]{wang2018execution}
Chenglong Wang, Kedar Tatwawadi, Marc Brockschmidt, Po-Sen Huang, Yi~Mao,
  Oleksandr Polozov, and Rishabh Singh. 2018.
\newblock \href {https://arxiv.org/abs/1807.03100} {Robust text-to-{SQL}
  generation with execution-guided decoding}.
\newblock \emph{arXiv preprint arXiv:1807.03100}.

\bibitem[{Yu et~al.(2018)Yu, Zhang, Yang, Yasunaga, Wang, Li, Ma, Li, Yao,
  Roman, Zhang, and Radev}]{yu2018spider}
Tao Yu, Rui Zhang, Kai Yang, Michihiro Yasunaga, Dongxu Wang, Zifan Li, James
  Ma, Irene Li, Qingning Yao, Shanelle Roman, Zilin Zhang, and Dragomir Radev.
  2018.
\newblock \href {https://doi.org/10.18653/v1/D18-1425} {{S}pider: A large-scale
  human-labeled dataset for complex and cross-domain semantic parsing and
  text-to-{SQL} task}.
\newblock In \emph{Proceedings of the 2018 Conference on Empirical Methods in
  Natural Language Processing}, pages 3911--3921, Brussels, Belgium.
  Association for Computational Linguistics.

\bibitem[{Zheng et~al.(2023)Zheng, Chiang, Sheng, Zhuang, Wu, Zhuang, Lin, Li,
  Li, Xing, Zhang, Gonzalez, and Stoica}]{zheng2023judging}
Lianmin Zheng, Wei-Lin Chiang, Ying Sheng, Siyuan Zhuang, Zhanghao Wu, Yonghao
  Zhuang, Zi~Lin, Zhuohan Li, Dacheng Li, Eric~P. Xing, Hao Zhang, Joseph~E.
  Gonzalez, and Ion Stoica. 2023.
\newblock \href {https://arxiv.org/abs/2306.05685} {Judging {LLM}-as-a-judge
  with {MT}-bench and chatbot arena}.
\newblock In \emph{Advances in Neural Information Processing Systems 36
  (NeurIPS 2023) Track on Datasets and Benchmarks}.

\bibitem[{Zhong et~al.(2020)Zhong, Yu, and Klein}]{zhong2020semantic}
Ruiqi Zhong, Tao Yu, and Dan Klein. 2020.
\newblock \href {https://doi.org/10.18653/v1/2020.emnlp-main.29} {Semantic
  evaluation for text-to-{SQL} with distilled test suites}.
\newblock In \emph{Proceedings of the 2020 Conference on Empirical Methods in
  Natural Language Processing (EMNLP)}, pages 396--411, Online. Association for
  Computational Linguistics.

\end{thebibliography}
